\documentclass{article}

\usepackage[preprint]{neurips_2026}
\usepackage{amsmath}
\usepackage[utf8]{inputenc} % allow utf-8 input
\usepackage[T1]{fontenc}    % use 8-bit T1 fonts
\usepackage{hyperref}       % hyperlinks
\usepackage{url}            % simple URL typesetting
\usepackage{booktabs}       % professional-quality tables
\usepackage{amsfonts,amsmath,amssymb}       % blackboard math symbols
\usepackage{nicefrac}       % compact symbols for 1/2, etc.
\usepackage{microtype}      % microtypography
\usepackage{xcolor}         % colors
\usepackage{capt-of}
\usepackage{graphicx}
\usepackage{subcaption}
\usepackage{cleveref}
\usepackage{xcolor}

\usepackage{todonotes}

\title{Muon in Vision Transformers: Optimizer–Recipe Interactions and Gradient Spectra}
\author{%
  Ben S. Southworth \\
  Los Alamos National Laboratories\\
  Los Alamos, NM 87545\\
  \texttt{southworth@lanl.gov} \\
  \And
  Shuai Jiang \\
  Sandia National Laboratories\\
  Albuquerque, NM 87123 \\
  \texttt{sjiang@sandia.gov} \\
  \And
  Daniel McBride \\
  Los Alamos National Laboratories\\
  Los Alamos, NM 87545\\
  \texttt{dmcbride@lanl.gov} \\
  \AND
  Eric C. Cyr \\
  Sandia National Laboratories\\
  Albuquerque, NM 87123 \\
  \texttt{eccyr@sandia.gov} \\
  \And
  Stephen Thomas \\
  Lehigh University \\
  Bethlehem, PA 18015\\
}

\begin{document}

\maketitle

\begin{abstract}
Muon is a recently developed matrix-aware optimizer that has shown strong results in transformer training, but its behavior in vision transformers (ViTs) is not yet well understood. 
We study Muon for ViT training, largely on ImageNet-100 and Pl@ntNet-300K, comparing against AdamW under standard vision recipes involving mixup, cutmix, smoothing, and random augmentation and erasing. 
Muon consistently outperforms AdamW, with especially large gains on long-tailed Pl@ntNet macro top-1. 
These gains are also recipe-dependent, where Muon benefits much more than AdamW from advanced and significant data augmentation techniques. 
To understand this interaction, we analyze the singular-value structure of matrix gradients throughout the ViT. 
Within Muon training runs, removing heavy data augmentation induces a late-training spectral concentration and mode collapse in gradient matrices, primarily in deep MLP-down blocks.  
Under a fixed ``full'' augmentation recipe, the clearest Muon--AdamW contrast appears instead in QKV gradients, where AdamW gradient energy remains concentrated in a much narrower basis while Muon spreads energy across substantially more singular modes. 
Muon in ViTs is therefore best understood as an optimizer–recipe interaction. Under a fixed recipe, Muon differs from AdamW most clearly in attention projections, where its gradients consist of a broader spectral basis. Within Muon, a full training recipe is important for preventing late spectral concentration and mode collapse in deep feedforward blocks.
We further demonstrate efficacy in training ViTs on image segmentation and masked autoencoder models, where Muon outperforms AdamW in all settings considered.
\end{abstract}

\section{Introduction}
\label{sec:intro}
Vision transformers (ViTs) are built from large learned linear maps. 
The dominant trainable blocks are the query/key/value (QKV) projections, the attention output projections, and the MLP projections \cite{dosovitskiy2021vit}. 
Despite this, ViTs are still typically trained with coordinate-wise optimizers such as AdamW \cite{loshchilov2019adamw}, which treat these matrices as collections of independent scalar parameters. 
Since most weights in a ViT are matrix-valued, a natural question is whether a matrix-aware optimizer can provide improved optimization over coordinate-wise methods. 

Muon is a recently developed matrix-aware optimizer for training neural networks
\cite{jordan2024muon,liu2025muonscalable}. 
Rather than applying an entry-wise adaptive update, Muon forms a momentum-like matrix update and approximately orthogonalizes it. 
In singular-value terms, this preserves the active singular directions of the update while suppressing singular-value anisotropy/ill-conditioning. 
This basic perspective has already motivated a growing literature around Muon and related matrix-aware or polar-factor methods, including scalable implementations, spectral interpretations, and alternative polar approximations \cite{jordan2024muon,liu2025muonscalable,amsel2025polarexpress,lau2025polargrad,chen2025spectralmuon,boreiko2025understandingmuon,riabinin2025gluon,dragutinovic2026simplicity, southworth2026beyond}. 
Most of this work, however, has focused only on language models or on optimizer analysis in the abstract. 
By comparison, Muon in ViTs or image foundation models has received minimal direct study.

This gap is somewhat surprising, as ViTs seem like a natural setting for Muon, and several related ideas have already come up in ViT literature. 
Orthogonal structure has been used as an architectural prior in ViTs \cite{wang2022ovit,huang2022orthogonaltransformer}, and approximately orthogonal parameterizations have also been used for efficient adaptation of pretrained ViTs \cite{yang2025orthogonalvitft}. 
Recent results indicate that Muon-like methods can perform well outside the language-model setting, including in long-tailed learning and vision-adjacent tasks \cite{anonymous2026longtailedmuon,wang2025tailend,chen2025muonad}, and at least one application-oriented paper that has identified Muon as being an effective optimizer for ViTs \cite{ancey2026fastposevit}. 
However, to the best of our knowledge the literature is missing a detailed ViT-focused study (or even demonstration) of when Muon outperforms standard AdamW % coordinate-wise optimizers, % we only compared against AdamW... technically
where in the architecture it affects training dynamics, and how this depends on standard vision training recipes.

This paper studies these questions on ViT classifiers for ImageNet-100 \cite{russakovsky2015imagenet} and Pl@ntNet-300K \cite{garcin2021plantnet300k}.\footnote{We note that Muon can also be applied to convolutional networks, but in our experience its gains are more modest there than in ViTs (see \Cref{app:plantnet:xception}).}
We compare Muon primarily against AdamW under standard ViT training recipes involving Mixup/CutMix and strong augmentation \cite{zhang2018mixup,yun2019cutmix,cubuk2020randaugment,zhong2020randomerasing}. 
Our first and main observation is that Muon works very well on these problems, and in our experiments substantially outperforms AdamW (\Cref{sec:main_results}). 
Our second point is more subtle, showing that the gain is not uniform across recipes. 
Muon benefits much more strongly from mixing-induced label smoothing than AdamW does. 
This is especially visible on long-tailed Pl@ntNet, where the improvement in macro top-1 is much larger than the improvement in standard top-1, and AdamW actually suffers from too strong of a data-augmentation recipe. This suggests that Muon depends strongly on the geometry of the training gradients induced by the data recipe.

\Cref{sec:muon_aug} and \Cref{sec:muon_adamw} explore this through a spectral analysis of ViT gradient matrices. 
Our first finding is a \emph{within-Muon} effect: removing strong augmentation when training with Muon causes gradient matrices to become spectrally concentrated late in training, most sharply in deep MLP-down blocks, coinciding with stagnation in validation metrics. 
Our second finding is an \emph{optimizer-contrastive} effect: under the same augmented recipe, the sharpest Muon--AdamW difference appears not in MLP-down blocks but in QKV gradients, where AdamW gradients remain concentrated in a few dominant singular directions throughout training while Muon spreads weight across substantially more from early on.

Muon in ViTs is therefore best understood as an optimizer–recipe interaction. Under a fixed recipe, Muon differs from AdamW most clearly in attention projections, where its gradients consist of a broader spectral basis. Within Muon, a full training recipe is important for preventing late spectral concentration and mode collapse in deep feedforward blocks. This perspective explains both why Muon is particularly effective in ViTs and why its gains are amplified by a standard strong vision recipe, explaining the results we see in Pl@ntNet.
Finally, \Cref{sec:beyond} demonstrates that this advantage extends to other learning tasks, with Muon outperforming AdamW in image segmentation \cite{xie2021segformer} and masked autoencoder training \cite{he2022masked}.

\paragraph{Contributions.}
The main contributions of the paper are as follows.
\begin{itemize}
    \item We present a systematic study of Muon for ViT training on ImageNet-100 and Pl@ntNet-300K, showing that Muon consistently outperforms AdamW. In addition, Muon benefits disproportionately from advanced data augmentation, especially mixup, cutmix, and label smoothing. 

    \item We introduce a spectral analysis of ViT matrix gradients based on cumulative singular-value energy. Within Muon, removing augmentation induces late-training spectral concentration, primarily in deep MLP-down blocks, and this transition coincides with stagnation in validation performance. 

    \item We perform an optimizer-controlled comparison under matched augmentation recipes and show that the AdamW QKV mappings are dominated by a small number of modes, whereas Muon rapidly spreads gradient energy across a broader set of singular directions.

    \item We demonstrate empirically that Muon outperforms AdamW as an optimizer for ViT image segmentation and masked autoencoder models. 
\end{itemize}

\section{Background: Muon and matrix-valued gradients in ViTs}
\label{sec:background}
\textbf{Muon as approximate polar-factor descent} --
Consider a matrix-valued parameter block $W_t \in \mathbb{R}^{m \times n}$ with gradient matrix $G_t$ at training step $t$. Muon forms a Nesterov momentum matrix state $M_t$ and updates the parameter by applying an approximate orthogonalization to this state \cite{jordan2024muon,liu2025muonscalable}:
\[
V_t = \beta V_{t-1} + G_t, \quad
M_t = G_t + \beta V_{t},
\quad
W_{t+1} = W_t - \eta\, \mathfrak{O}(M_t),
\]
where $\mathfrak{O}(\cdot)$ denotes the orthogonalization operator used by Muon. In practice, Muon specifically approximates the orthogonal polar factor. If
$
M_t = U \Sigma V^\top
$
is the singular value decomposition (SVD) of $M_t$, then
\[
\mathrm{polar}(M_t) = U V^\top.
\]
Thus, the polar-factor update removes the singular values of $M_t$ while preserving its left and right singular directions meaning large singular values no longer dominate the update purely because of scale, while weaker but still active singular directions become relatively more influential. 

Muon therefore acts directly on the spectral anisotropy/ill-conditioning of a matrix update rather than on its entries in isolation. 
For gradient matrices considered in this paper, $G_t$ and $M_t$ consistently have singular values spanning 2--4 orders of magnitude, with relatively sharp decay at the beginning (see Appendix \ref{app:spectral_details}), indicating that without normalization, updates would be dominated by a small number of directions.
% This is closely related in spirit to whitening. A whitening map suppresses correlation between basis functions, and, generalizing to a matrix, setting corresponds to ensuring the whitened operator is orthogonal (see \Cref{app:exact_whitening}). \todo{I'm a little lost on exact whitening}
%Thus Muon targets a polar whitening of the update operator itself, retaining the singular subspaces and suppressing singular-value imbalance. 
In practice Muon uses efficient iterative approximations to the polar factor, typically Newton--Schulz style iterations and their refinements, allowing for competitiveness in LLM pretraining \cite{jordan2024muon,amsel2025polarexpress,lau2025polargrad}. 

We focus on the standard practical Muon variant used in our experiments, with optimized coefficients from \cite{amsel2025polarexpress}. Muon introduces additional computational cost over AdamW, but in the ViT setting Muon's relative overhead cost is smaller than in LLMs, and in all of our tests modest enough (see Appendix \ref{app:cost}) that the accuracy gains provided by Muon are significant when normalized in wall-clock terms. A broader perspective on Muon as a whitening approximation, together with several nearby variants that we tested is provided in Appendix \ref{app:whitening_general}.

\textbf{Vision transformers} --
Despite the vast and growing number of ViT applications, their computational core remains the same: dense linear maps \cite{dosovitskiy2021vit}. 
Within any given transformer block, the QKV, attention output, and MLP up/down projections drive the majority of trainable matrix computations. 
Because these parameter types are sensitive to singular directions, spectral concentration, and operator-valued update geometry \cite{hu2024specformer,murata2026svd}, ViTs provide a suitable environment for analyzing Muon. 
Furthermore, this setting isolates how Muon behaves across different large linear operators. 
We find that the optimizer's effects are highly localized: the sharpest contrast between Muon and AdamW manifests in QKV gradients, while deep MLP-down blocks exhibit the strongest augmentation-dependent effects within Muon itself.

\textbf{Mixing-side smoothing and matrix gradients} --
Modern vision recipes rely heavily on data-side smoothing and regularization, including label smoothing, Mixup, CutMix, RandAugment, and random erasing \cite{zhang2018mixup,yun2019cutmix,cubuk2020randaugment,zhong2020randomerasing}. 
In addition to broadly improved generalization, for this paper an important nuance is that they change the matrix gradients seen by the optimizer. A useful heuristic comes from the outer-product structure of linear-layer gradients. For a single example, the gradient of a weight matrix $W$ typically has the form
$
\nabla_W \ell = \boldsymbol{\delta}\mathbf{x}^\top,
$
where $\mathbf{x}$ is the layer input and $\boldsymbol{\delta}$ is the backpropagated upstream signal \cite{martens2015optimizing}. A minibatch gradient is therefore a sum of outer products,
\[
G = \sum_{i=1}^{B} \boldsymbol{\delta}_i \mathbf{x}_i^\top.
\]
Under Mixup, however, the model is trained on convex combinations of examples and targets,
$
\tilde{\mathbf{x}} = \lambda \mathbf{x}_i + (1-\lambda)\mathbf{x}_j$, and $\tilde{\mathbf{y}} = \lambda \mathbf{y}_i + (1-\lambda)\mathbf{y}_j. $
Under a local linearization, the corresponding gradient contribution takes the schematic form
\[
\tilde{\boldsymbol{\delta}}\, \tilde{\mathbf{x}}^\top
\approx
(\lambda \boldsymbol{\delta}_i + (1-\lambda)\boldsymbol{\delta}_j)
(\lambda \mathbf{x}_i + (1-\lambda)\mathbf{x}_j)^\top,
\]
which expands into both self terms such as $\boldsymbol{\delta}_i \mathbf{x}_i^\top$ and cross terms such as $\boldsymbol{\delta}_i \mathbf{x}_j^\top$ and $\boldsymbol{\delta}_j \mathbf{x}_i^\top$. CutMix induces a closely related effect, though with spatially localized rather than global mixing. In both cases, the net result is that mixing-based smoothing can enrich the \emph{span} of minibatch gradient matrices and reduce concentration in a small number of dominant singular directions.
Thankfully, this is the exact regime where one expects Muon's update to be effective.
If the gradient matrix contains support across many meaningful directions, then suppressing singular-value anisotropy can redistribute the update across a broader active subspace. 
If, on the other hand, the gradient matrix collapses into a narrow basis, then the effective subspace available to Muon narrows accordingly.
A different perspective is also considered in Appendix~\ref{app:linear_model} where augmentations as linear maps are considered. 

\section{Muon benefits from mixing, smoothing, and augmentation}
\label{sec:main_results}

We begin with the empirical phenomenon that motivates the rest of the paper. Muon substantially outperforms AdamW on both ImageNet-100 and Pl@ntNet-300K when training ViT-B/16. However, the gain is not uniform across training recipes. It is largest under a full training recipe incorporating more data mixing, smoothing, and augmentation, and the success of Muon is especially pronounced on long-tailed Pl@ntNet macro top-1.

\textbf{Experimental setting} -- 
Our primary experiments train ViT-B/16 on ImageNet-100 and Pl@ntNet-300K using either AdamW or Muon.
%\todo{I'm currently doing doing the plantnet; i think the commented out blurb below is useful to put in appendix (even if it's the default from pytorch stuff apparently)}
% All runs use the same baseline spatial preprocessing: random resized cropping with minimum scale $0.08$ and random horizontal flipping with probability $0.5$. Evaluation uses crop ratio $0.875$. Thus, the weakest setting considered here is best interpreted as baseline spatial preprocessing only, rather than as the complete absence of augmentation.\todo{This seem like too many numbers, should reword and move more details to appendix in tables}
All runs use the same baseline spatial preprocessing of random resized cropping and horizontal flipping. 
%% BS : no on itemize
% \todo{This seemed like too many numbers, should reword and move more details to appendix in tables}
% On top of this baseline, we vary two additional recipe components:
% \begin{itemize}
%     \item \emph{Strong augmentation}: refers to RandAugment and random erasing.
%     \item \emph{Mixing-side smoothing}: refers to the coupled use of Mixup, CutMix, and label smoothing. Because these three are enabled or disabled together in our experiments, we treat them as a single component in the main study.
% \end{itemize}
%
% We then consider four representative settings:\todo{I made it an itemized list... but not sure if i like it}
%  \begin{itemize}
%     \item \emph{Full} (both components enabled)
%     \item \emph{No Rand} (strong augmentation removed)
%     \item \emph{No Mix} (mixing-side smoothing removed)
%     \item \emph{No Mix/No Rand} (both removed)
% \end{itemize} 
% On ImageNet-100 we report validation top-1 accuracy. 
% On Pl@ntNet-300K we report both top-1 and macro top-1, since the latter is more informative in the long-tailed setting. 
% % All main experiments use batch size 512 and run for 15{,}000 training steps. 
% Complete hyperparameter details are given in Appendix \ref{app:hyperparams}.
% % and analogous results for batch size 128 and comparable 60{,}000 steps in \todo{ref}.
On top of this baseline, we vary two additional recipe components. We use \emph{strong augmentation} to refer to RandAugment and random erasing, and \emph{mixing-side smoothing} to refer to the coupled use of Mixup, CutMix, and label smoothing. Because Mixup, CutMix, and label smoothing are enabled or disabled together in our experiments, we treat them as a single component in the main study. 
We then consider four representative settings: \emph{Full} (both components enabled), \emph{No Rand} (strong augmentation removed), \emph{No Mix} (mixing-side smoothing removed), and \emph{No Mix/No Rand} (both removed). On ImageNet-100 we report validation top-1 accuracy. On Pl@ntNet-300K we report both top-1 and macro top-1, since the latter is more informative in the long-tailed setting. All main experiments use batch size 512 and run for 15{,}000 training steps. 
% Complete hyperparameter and analogous results for batch size 128 and comparable 60,000 steps in Appendix~\ref{app:plantnet}.
Complete hyperparameters and further ablation studies are provided in Appendix~\ref{app:plantnet}.

\begin{figure*}[t]
\centering
\begin{minipage}[t]{0.44\textwidth}
    \vspace{-1.5ex}
    \centering
    \includegraphics[width=2.5in]{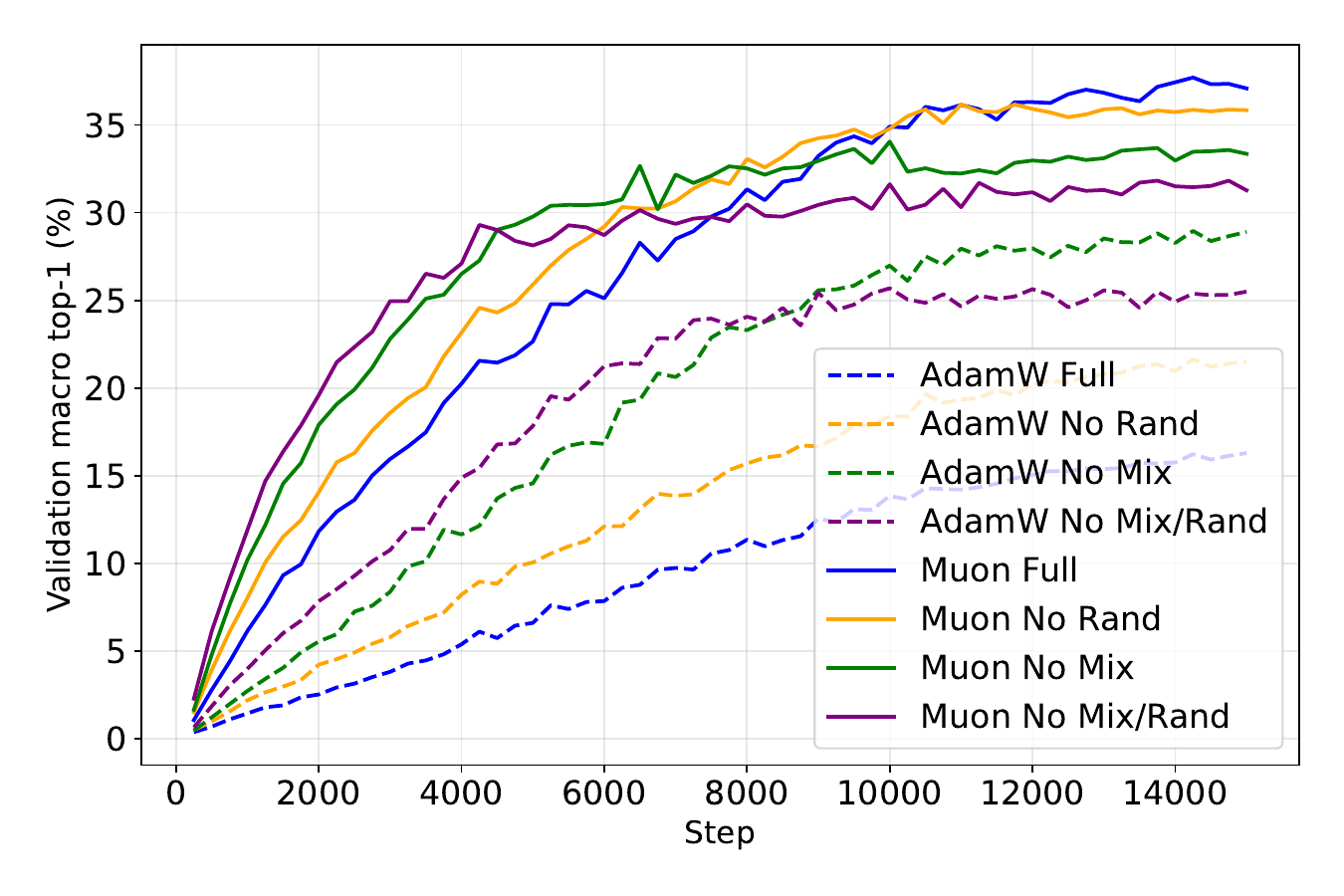}
    
    \vspace{-1.5ex}
    \caption{Validation macro top-1 on Pl@ntNet-300K for AdamW and Muon under representative training recipes. Muon benefits much more strongly from the full recipe than AdamW, and the gap is sharpest on the long-tailed macro metric.}
    \label{fig:plantnet_macro_top1}
\end{minipage}
\hfill
\begin{minipage}[t]{0.55\textwidth}
    \vspace{0pt}
    \centering
    \small
    \setlength{\tabcolsep}{3.5pt}
    \begin{tabular}{llccc}
    \toprule
    Optimizer & Recipe & IN100 & Pl@ntNet & Macro \\
    \midrule
    AdamW & Full        & 66.18 & 63.58 & 16.30 \\
    AdamW & No Rand     & 67.00 & 69.00 & 18.28 \\
    AdamW & No Mix      & 65.20 & 71.04 & 28.90 \\
    AdamW & No Mix/No Rand & 59.48 & 66.70 & 25.50 \\
    \midrule
    Muon  & Full        & \textbf{81.20} & \textbf{80.39} & \textbf{37.70} \\
    Muon  & No Rand     & 78.78 & 79.43 & 35.84 \\
    Muon  & No Mix      & 74.50 & 76.23 & 33.57 \\
    Muon  & No Mix/No Rand & 68.18 & 73.93 & 31.52 \\
    \bottomrule
    \end{tabular}
    \captionof{table}{Final validation accuracy under representative training recipes. Muon outperforms AdamW across all settings, with the strongest gains under the full recipe and the sharpest advantage on Pl@ntNet macro top-1.}
    \label{tab:main_results}
\end{minipage}
    \vspace{-2ex}
\end{figure*}

\textbf{Muon outperforms AdamW across datasets and recipes} -- ~\Cref{fig:plantnet_macro_top1} and  Table~\ref{tab:main_results} show final validation accuracy across representative recipe settings, and the specific Pl@ntNet macro top-1 validation evolution. 
First, we emphasize that Muon outperforms AdamW in all of the settings we tested, across training recipe, dataset, and validation measure (including validation loss not shown here).
Second, note that on Pl@ntNet, the clearest gain is in macro top-1. Under the full recipe, Muon achieves a $+21.40$ point advantage over AdamW. Even under weaker recipes, Muon's lead remains significant, yielding improvements of $+17.56$ points under \emph{No Rand}, $+4.67$ under \emph{No Mix}, and $+6.02$ under \emph{No Mix/No Rand}.
% Second, note that on Pl@ntNet, the clearest gain is not in standard top-1 but in macro top-1. Under the full recipe, Muon reaches 37.70 macro top-1 while AdamW reaches only 16.30. 
% Even under weaker recipes, Muon remains substantially ahead: 35.84 vs.\ 18.28 under \emph{No Rand}, 33.57 vs.\ 28.90 under \emph{No Mix}, and 31.52 vs.\ 25.50 under \emph{No Mix/No Rand}. 
% This is difficult to dismiss as a simple speed-of-optimization effect. 
As suggested by \cite{anonymous2026longtailedmuon}, the strongest advantage appears on the metric that is most sensitive to long-tail performance.

Beyond the generally strong performance of Muon, the further important point of this section is that Muon benefits distinctly from the training recipe. On ImageNet-100, AdamW varies only modestly across the stronger recipe settings, while Muon gains $13$ points in validation top-1. The largest gain in Muon is from incorporating  mixing-side smoothing rather than strong augmentation. The same pattern is even clearer on Pl@ntNet. AdamW actually suffers from the \emph{Full} recipe, wherein \emph{removing} mixing-side smoothing improves AdamW from 63.58 / 16.30 to 71.04 / 28.90 in top-1 / macro top-1. Muon yields the \emph{opposite behavior}. Its best performance is obtained under the \emph{Full} recipe, and performance degrades systematically as recipe components are removed. 
Two points follow: first, Muon's efficacy depends strongly on the data recipe. 
Second, the component that matters most appears to be mixing-side smoothing rather than stronger image augmentation.

\section{Why mixing-side smoothing and strong augmentation matter for Muon}
\label{sec:muon_aug}

\Cref{sec:main_results} showed that Muon benefits disproportionately from mixing-side smoothing and strong augmentation. 
We now study the singular structure in matrix gradients arising during training Pl@ntNet-300K using the full recipe versus the \emph{No Mix/No Rand} setting, reflecting the removal of both mixing-side smoothing and strong augmentation.

\textbf{Spectral summaries of matrix gradients} --
Let $G \in \mathbb{R}^{m \times n}$ be a matrix-valued gradient block with singular values
$
\sigma_1 \ge \sigma_2 \ge \cdots \ge \sigma_r \ge 0,$ where $r = \min(m,n).$
We summarize its singular structure through the cumulative energy curve
\[
C(\mu) =
\frac{\sum_{i=1}^{\lfloor \mu r \rfloor} \sigma_i^2}
{\sum_{i=1}^{r} \sigma_i^2},
\qquad
\mu \in [0,1],
\]
which represents the fraction of total spectral energy captured by the top $\mu$-fraction of singular directions. 
A steep curve indicates concentration in a small number of dominant modes while a flatter curve indicates broader spectral support. For compact comparison we use the energy-quantile rank
\[
\mu_p = \inf\{u \in [0,1] : C(\mu) \ge p\},
\]
that is, the smallest normalized rank required to explain a fraction $p$ of total energy. Larger values of $\mu_p$ correspond to broader spectra. 

In the heatmaps below we compare runs through \emph{ratios} of these quantities across matched linear blocks and training steps. 
Ratios greater than one therefore indicate that one run requires more singular directions than the other to explain the same amount of gradient energy.

\begin{figure}[t]
    \centering
    \includegraphics[width=0.9\columnwidth]{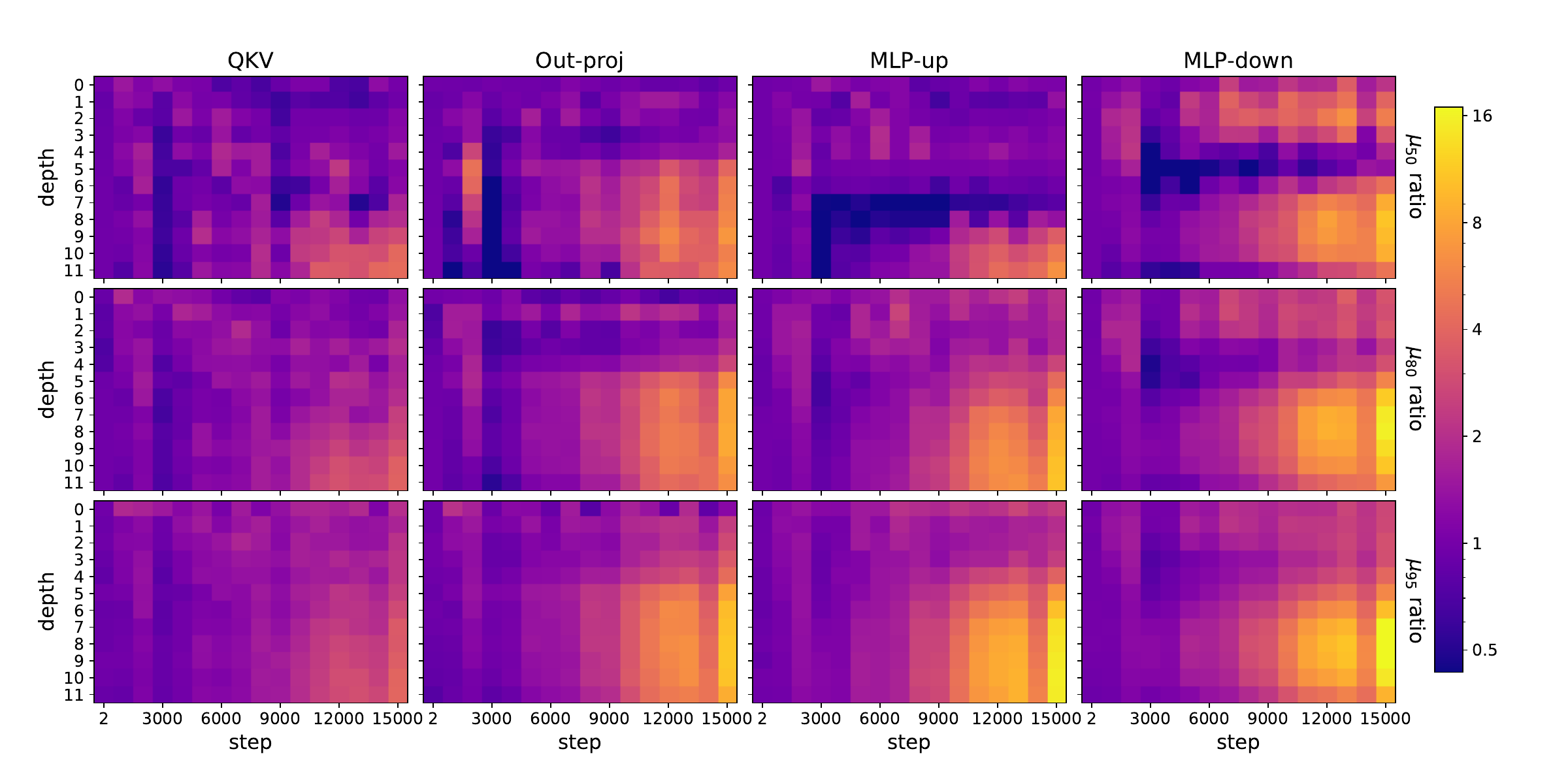}
    \caption{Energy-quantile rank ratio for Muon trained with the full recipe over Muon trained with \emph{No Mix/No Rand} for gradient matrices $G$. Values greater than one indicate broader spectra under the full recipe. The strongest late-training effect is concentrated in deep MLP-down blocks.}
    \label{fig:muon_aug_atlas}
    % \vspace{-4ex} % looks weird 
\end{figure}
\textbf{Removing mixing-side smoothing and strong augmentation induces late spectral concentration} --
\Cref{fig:muon_aug_atlas} plots the ratio of energy quartiles for Muon trained under the \emph{Full} recipe to Muon trained under \emph{No Mix/No Rand}. Broadly, under the \emph{Full} recipe Muon requires more singular directions to explain the same amount of gradient energy, indicating broader spectral support as suggested by the gradient analysis in \Cref{sec:background}. 
The ratio shows large changes late in training, near the same stage at which the weaker-recipe Muon run begins to stall in validation top-1 and macro top-1 (e.g. the progress in \Cref{fig:plantnet_macro_top1} largely stagnates around step 9k).
% Removing mixing-side smoothing and strong augmentation has the strongest effect late in training -- differences are not present at the start of or early in optimization, but emerge later, near the same stage at which the weaker-recipe Muon run begins to stall in validation top-1 and macro top-1. 
The effects of the full recipe are also depth-localized. 
The largest changes occur in middle-to-deep transformer blocks rather than uniformly throughout the network. 
Lastly, full-recipe effects are specific to the class of matrix parameters (QKV, Out-projection, and MLP-up/down).
The strongest and clearest ratios appear in MLP-down, with slightly weaker but still visible structure in MLP-up and much smaller changes in QKV blocks.

\begin{figure*}[t]
    \centering
    \begin{subfigure}[t]{0.48\textwidth}
        \centering
        \includegraphics[width=\linewidth]{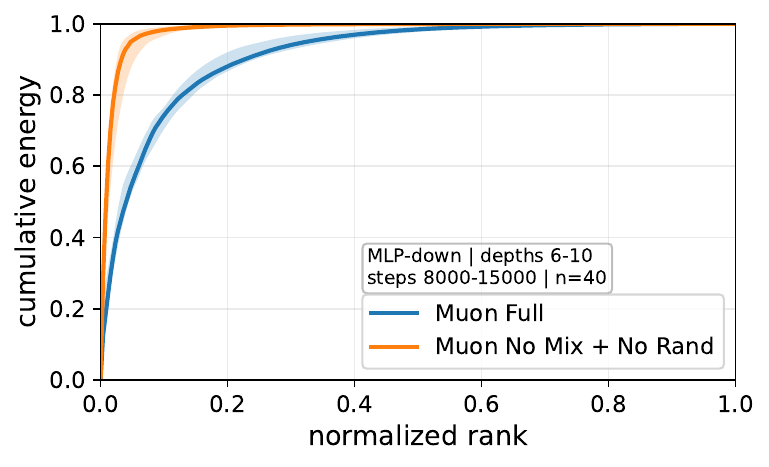}
        \caption{Within Muon runs: cumulative spectral-energy summaries for deep MLP-down blocks, comparing the full recipe against \emph{No Mix/No Rand} over the late-training window highlighted by the heatmap. Solid lines show the median across selected block snapshots and shaded regions show the interquartile range.}
        \label{fig:curve_summary_muon_aug}
    \end{subfigure}
    \hfill
    \begin{subfigure}[t]{0.48\textwidth}
        \centering
        \includegraphics[width=\linewidth]{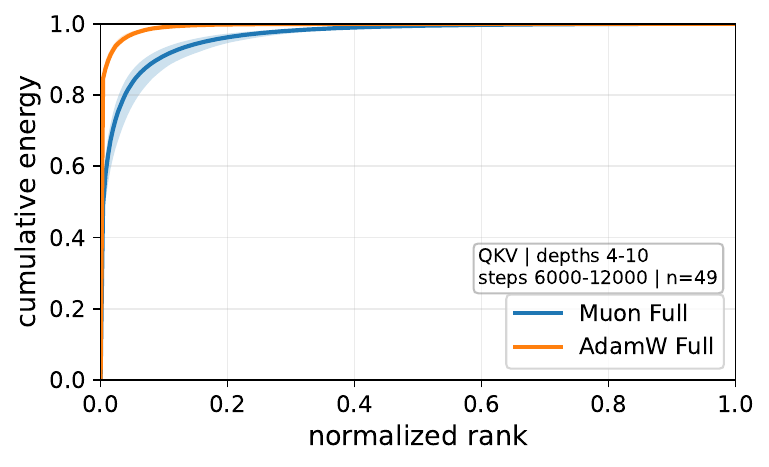}
        \caption{Muon vs.\ AdamW under the full recipe: cumulative spectral-energy summaries for QKV blocks over the representative depth and step range highlighted by the optimizer-comparison heatmap. Solid lines show the median across selected block snapshots and shaded regions show the interquartile range.}
        \label{fig:curve_summary_muon_adamw}
    \end{subfigure}
    \caption{Representative cumulative spectral-energy summaries. Left: within Muon, removing mixing-side smoothing and strong augmentation yields steeper late-training MLP-down curves, indicating concentration in fewer singular directions. Right: under the full recipe, AdamW exhibits markedly steeper QKV curves than Muon, indicating a narrower active gradient basis.}
    \label{fig:curve_summary_both}
    % \vspace{-2ex}
\end{figure*}

Altogether, in many deep blocks late in training, we consistently see a significant $10-16\times$ broader energy support in the gradient matrices using the \emph{Full} training recipe compared with \emph{No Mix/No Rand}. 
The differences arise because with \emph{No Mix/No Rand}, the gradient spectra become substantially more concentrated, which is shown directly in representative cumulative-energy summaries in~\Cref{fig:curve_summary_muon_aug}. 
Aggregating over the late-training deep MLP-down regime highlighted by the heatmap, the \emph{Full}-recipe run exhibits consistently flatter curves than the \emph{No Mix/No Rand} run. Full spectra aggregated across all operator can be seen in Appendix \ref{app:spectral_details}, and although the gradient matrices are not truly low-rank, they are increasingly dominated by a small set of modes. Analogous plots are also provided for momentum matrices in Appendix \ref{app:spectral_details}, but the ratios there are much closer to unity and less structure because the slow decay and averaging in momentum matrices obscures the distinct gradient behavior that arises with different training recipes.
The above observations provide a likely explanation for the recipe dependence in~\Cref{sec:main_results}. Mixing-side smoothing and strong augmentation maintain broader gradient spectral support late in training, particularly in deep MLP-down blocks, while under \emph{No Mix/No Rand} this support collapses precisely when validation performance stalls. 
This matters specifically for Muon because it suppresses singular-value anisotropy while preserving singular directions; a broad active subspace in the gradient matrix is therefore a prerequisite for redistributing updates across many meaningful modes. 
% When the gradient collapses into a narrow basis, so does Muon's effective update subspace.
Analogous results for smaller batch size 128 are shown in Appendix \ref{app:bsize128}.

% \textbf{Interpretation} --
% Together, these plots suggest that the \emph{Full} recipe causes  redistribution of gradient energy away from a small dominant basis to a broader set of singular directions. 
% These observations explain the recipe dependence from ~\Cref{sec:main_results}. Mixing-side smoothing and strong augmentation actively change the matrix gradients that Muon receives, and yield support across a broader set of singular directions. Under \emph{No Mix/No Rand}, this support collapses late in training, primarily in deep MLP-down, MLP-up, and Out-projection blocks. This concentration is closely associated with the point at which validation performance of Muon stops improving (\Cref{fig:plantnet_macro_top1}), while the broader spectral support of gradient matrices in the \emph{Full} recipe enables continued improvement. This is consistent with the update geometry of Muon. Muon acts by suppressing singular-value anisotropy while preserving singular directions. If the gradient matrix contains a broad active subspace, then Muon can redistribute the update across many meaningful modes. If the gradient matrix collapses into a narrow basis, then the effective subspace available to Muon narrows as well. From this viewpoint, mixing-side smoothing helps Muon by maintaining matrix-gradient richness late in training. Additional analyses, including the corresponding matrix state $M$ and batch size robustness, are reported in the appendix.\todo{which appendix?}

\section{How Muon differs from AdamW}
\label{sec:muon_adamw}

\Cref{sec:muon_aug} identified a within-Muon runs effect: removing mixing-side smoothing and strong augmentation induces late spectral concentration in deep MLP-down blocks. We now consider how Muon and AdamW differ, and demonstrate similar spectral differences but in QKV gradients when training Pl@ntNet-300K. To isolate optimizer effects from recipe effects, we compare Muon and AdamW under the same \emph{Full} training recipe. As in ~\Cref{sec:muon_aug}, we analyze gradient matrices $G$ across block families and depths using cumulative spectral energy and energy-quantile rank (although AdamW does not use matrix structure in its optimization, the relevant operators can still be constructed and analyzed). Larger quantile rank again indicates broader spectral support, and ratios greater than one indicate that Muon requires more singular directions than AdamW to explain the same amount of gradient energy. This comparison answers a different question from the one in ~\Cref{sec:muon_aug}. 
There, the issue was how and where mixing-side smoothing and strong augmentation prevents late spectral collapse of gradient matrices inside Muon training. 
Here, the question is where Muon induces distinct evolution of gradient geometry relative to the AdamW coordinate-wise optimizer.

\begin{figure}[t]
    \centering
    \includegraphics[width=0.9\columnwidth]{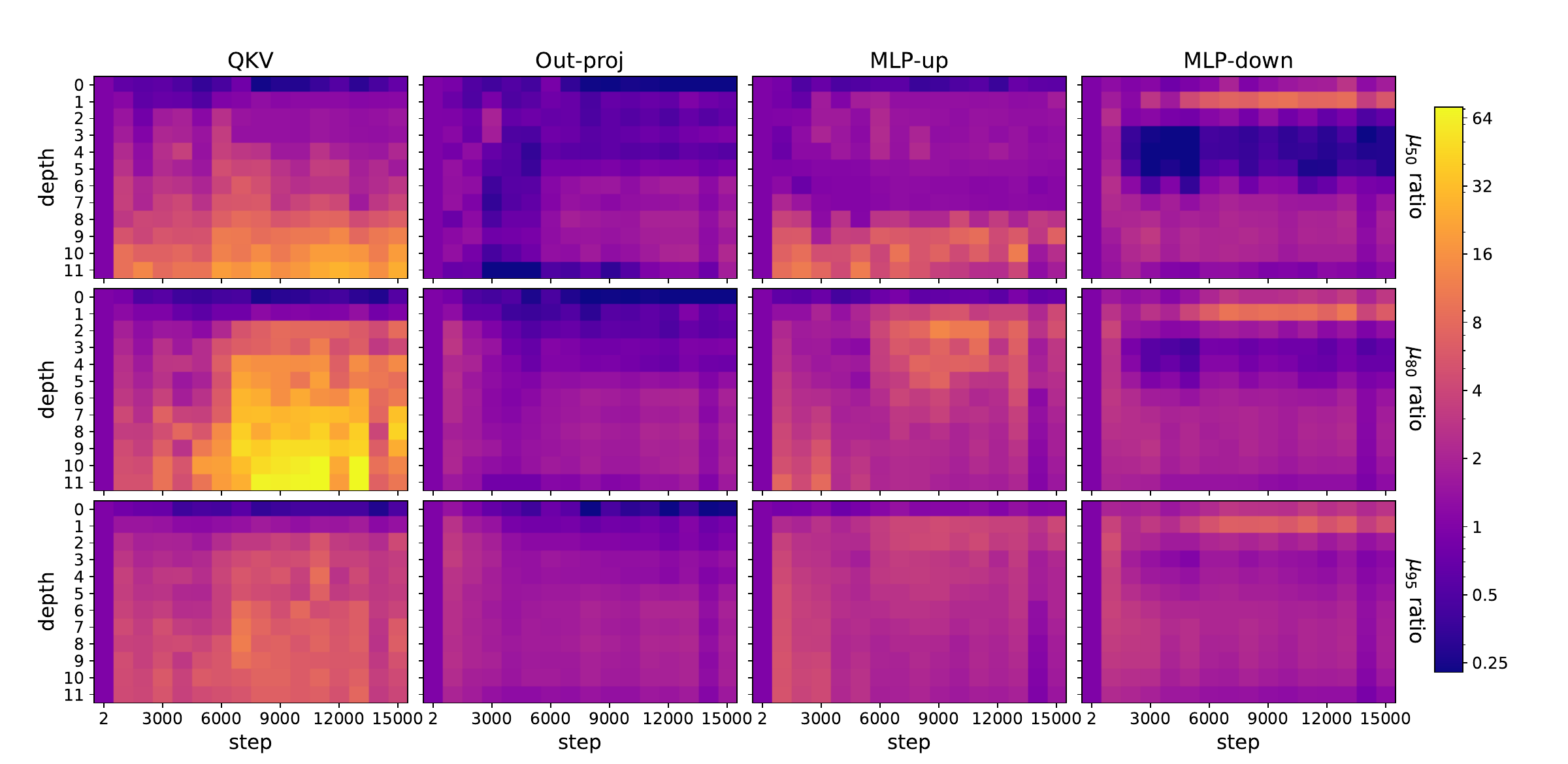}
    \caption{Energy-quantile rank ratios of Muon over AdamW for gradient matrices $G$ under the full recipe. Values greater than one indicate broader spectra under Muon. The clearest optimizer-level contrast appears in QKV gradients.} 
    \label{fig:muon_adamw_atlas}
\end{figure}

\textbf{Muon broadens QKV gradient support relative to AdamW} --
\Cref{fig:muon_adamw_atlas} shows the cumulative energy density quartile rank ratios of Muon to AdamW. In contrast to within-Muon results from \Cref{sec:muon_aug}, here the largest ratios arise in QKV matrices. Across depths and training steps, Muon consistently requires \emph{significantly} more singular directions than AdamW to explain the same fraction of QKV gradient energy. The effects of Muon appear early and persist through training rather than emerging only late, while deep in the network and later in training the ratio reaches as high as $64\times$. By contrast, the differences in output projections and MLP blocks are smaller and less spatially coherent, although Muon gradient matrices have larger energy quartiles in most cases. Representative cumulative-energy summaries in ~\Cref{fig:curve_summary_muon_adamw} make the same distinction more directly, and demonstrate that numerically AdamW QKV gradient matrices are dominated by a few singular modes. There we sum over a set of 49 QKV gradient matrices in the regime with large ratios in ~\Cref{fig:muon_adamw_atlas}, and it is clear that AdamW places a much larger fraction of energy in the leading few singular directions. Muon also has several dominant modes, but spreads out significantly more than AdamW.

\textbf{Interpretation} --
These results suggest that the strongest optimizer-specific \emph{Muon-vs-AdamW} effect in ViTs under a fixed recipe is in QKV attention input projections (although all transformer blocks do maintain meaningfully broader energy spectra when trained with Muon compared with AdamW), with AdamW QKV gradients being substantially more mode-concentrated than Muon gradients. This result is complementary to the augmentation result in ~\Cref{sec:muon_aug}, where the strongest \emph{within-Muon} effect of removing augmentation was a late concentration in deep MLP-down blocks, with secondary results in MLP-up and Out-projection blocks. Overall, this suggests two distinct layers of the mechanism. Muon differs from AdamW most directly in attention projections, while mixing-side smoothing and strong augmentation is most important for preventing late collapse of gradient support in deep feedforward blocks once Muon is already in use. Although AdamW lags behind Muon in validation performance at matched training steps (\Cref{fig:plantnet_macro_top1}), the same broad behavior of spectral concentration is visible when comparing runs at roughly matched validation progress as well.

\section{Beyond image classification}\label{sec:beyond}

Results so far have focused on image classification. We now briefly demonstrate that (i) Muon is an effective choice of optimizer in other ViT-based vision settings, outperforming AdamW in all cases, and (ii) our observations regarding optimizer-recipe interaction appear to persist in other settings.

\textbf{Image segmentation:} Evaluating AdamW and Muon in a transformer-based dense-prediction setting using a compact SegFormer model \cite{xie2021segformer} on LoveDA semantic segmentation \cite{wang2021loveda} yielded results consistent with the ViT-B/16 experiments. The comparison uses a matched design with two optimizers, four augmentation regimes in increasing order of augmentation strength (\texttt{noaug}, \texttt{base}, \texttt{geo}, \texttt{geophoto}), and three seeds; exact regime definitions are given in appendix \ref{app:image_segmentation}. As the primary validation metric we use mean intersection-over-union (mIoU), the standard class-balanced figure of merit for semantic segmentation \cite{long2015fully}. We also report overall accuracy (OA) and Cohen's kappa as supporting metrics.

Across all four augmentation regimes, Muon achieves higher terminal validation mIoU than AdamW, and OA and Kappa show the same overall pattern, as summarized in Table~\ref{tab:seg-beyond-cls-terminal}. The heaviest augmentation regime is especially informative: AdamW degrades from the geometric regime to the heavier geometric-plus-photometric regime ($0.232$ to $0.227$ for AdamW versus $0.267$ to $0.280$ for Muon), whereas Muon remains comparatively robust. This suggests that the optimizer--recipe interaction perspective extends beyond image classification and into transformer-based semantic segmentation. Validation curves, additional plots, and implementation details are deferred to appendix \ref{app:image_segmentation}.

\begin{table}[t]
    \centering
    \caption{Terminal validation results on LoveDA semantic segmentation for AdamW and Muon across augmentation regimes, reported as means over three seeds. Within each metric pair, the better value is shown in bold.}
    \label{tab:seg-beyond-cls-terminal}
    \begin{tabular}{lcccccc}
        \toprule
        Regime
        & \multicolumn{2}{c}{mIoU}
        & \multicolumn{2}{c}{OA}
        & \multicolumn{2}{c}{Kappa} \\
        \cmidrule(lr){2-3}
        \cmidrule(lr){4-5}
        \cmidrule(lr){6-7}
        & AdamW & Muon & AdamW & Muon & AdamW & Muon \\
        \midrule
        noaug    & 0.222 & \textbf{0.266} & 0.494 & \textbf{0.522} & 0.303 & \textbf{0.353} \\
        base     & 0.235 & \textbf{0.260} & 0.498 & \textbf{0.516} & 0.319 & \textbf{0.348} \\
        geo      & 0.232 & \textbf{0.267} & 0.499 & \textbf{0.526} & 0.317 & \textbf{0.360} \\
        geophoto & 0.227 & \textbf{0.280} & 0.513 & \textbf{0.556} & 0.322 & \textbf{0.385} \\
        \bottomrule
    \end{tabular}
\end{table}

\textbf{Foundation model:} 
We evaluate Muon in a self-supervised pretraining (PT) using masked autoencoding (MAE) \cite{he2022masked} trained on ImageNet-1k.
One can interpret the masking of MAE as a heavy form of data augmentation, but the lack of labels means that neither the above analysis of Mixup nor that of Appendix~\ref{app:linear_model} applies.
This therefore serves as a stress test for Muon outside the supervised classification setting.
However, we still do observe benefits from using Muon as an optimizer, consistent with results in LLM pretraining.

\begin{figure*}[t]
    \centering
    % Left Side: Table
    \begin{minipage}[c]{0.52\textwidth}
        \centering
        % \resizebox forces the table to scale to the width of the minipage
        \resizebox{\linewidth}{!}{%
        \begin{tabular}{lcccc}
            \toprule
            & \multicolumn{4}{c}{Top-1 Acc.} \\
            \cmidrule(lr){2-5}
            Model
            & \multicolumn{2}{c}{Pretrain: AdamW}
            & \multicolumn{2}{c}{Pretrain: Muon} \\
            \cmidrule(lr){2-3}
            \cmidrule(lr){4-5}
            & FT: AdamW & FT: Muon & FT: AdamW & FT: Muon \\
            \midrule
            ViT-B & 81.23 & {81.29} & 81.27 & \textbf{81.49} \\ % Chkpt 200, epoch 30 for now # Double check muon muon for run 
            ViT-L & 84.03 & 84.09  & 84.24 & \textbf{84.56} \\ % Checkpt 200, epoch 20 for now, 
            \bottomrule
        \end{tabular}%
        } % <-- end of resizebox
    \end{minipage}\hfill
    % Right Side: Figure
    \begin{minipage}[c]{0.45\textwidth}
        \centering
        \includegraphics[width=\linewidth]{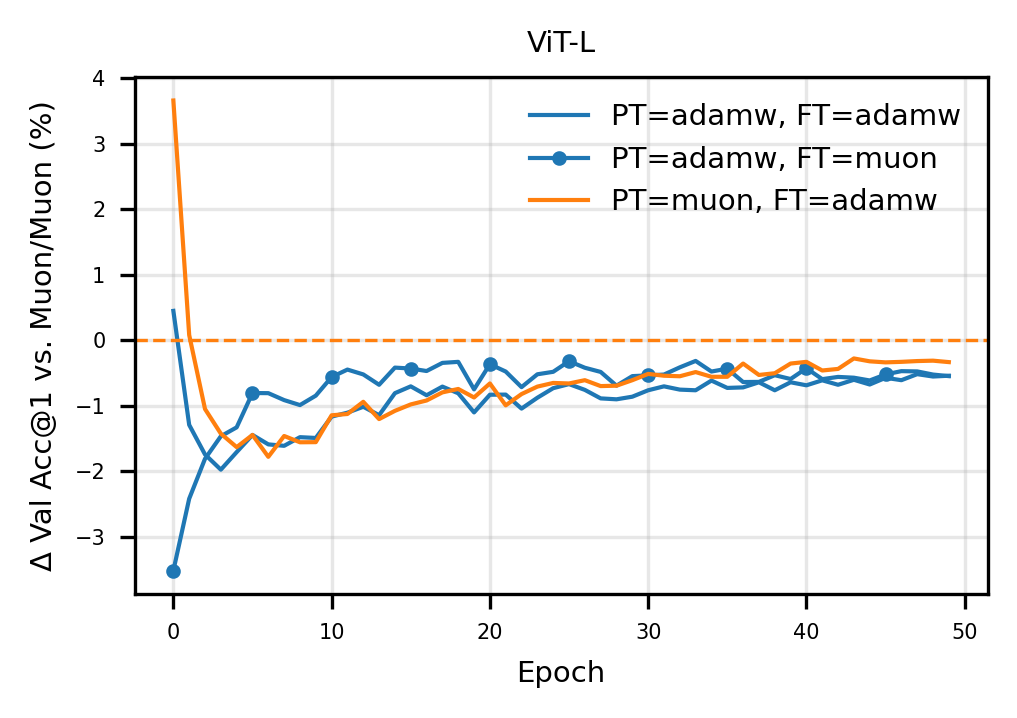}
    \end{minipage}
    
    % \vspace{0.2cm} % Small vertical buffer before the caption
    
    \caption{{Ablation of optimizer choice (AdamW vs.\ Muon) during MAE pretraining and finetuning.} \textbf{Left:} Best validation Acc@1 on ImageNet-1k across model sizes and optimizer combinations. \textbf{Right:} Validation Acc@1 relative to the Muon/Muon baseline for ViT-L. Utilizing Muon for both phases shows demonstrable benefits at any phase of FT.}
    \label{fig:finetune_ablation_combined}
\end{figure*}

\begin{figure}[t]                 
      \centering                            
      \includegraphics[width=.9\textwidth]{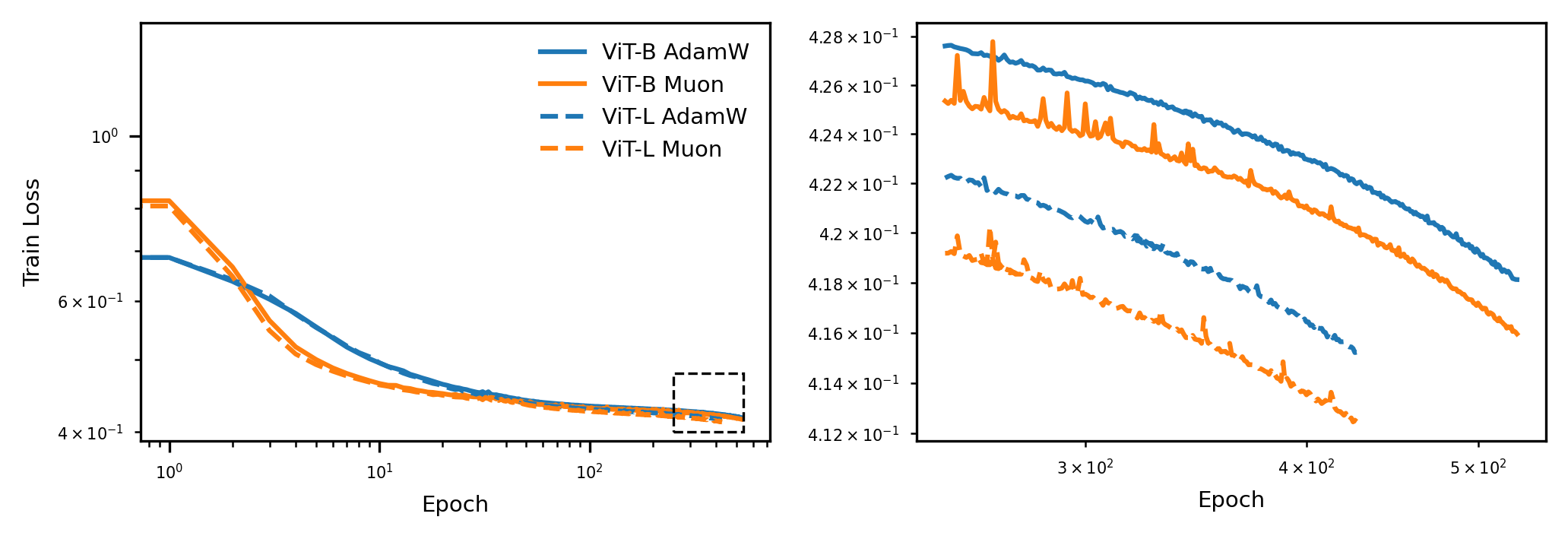}
      \caption{Pre-training loss curves on ImageNet for ViT-B and ViT-L under AdamW and Muon optimizers at same learning rate. \textbf{Left:} full training trajectories. \textbf{Right:} zoomed view of later epochs, revealing that Muon consistently has lower loss than AdamW.}
      \label{fig:mae_pretrain_losses}
  \end{figure}

In \Cref{fig:mae_pretrain_losses}, we plot the pre-training loss for ViT-B and ViT-L under a simple swap from AdamW to Muon, keeping \emph{all} hyperparameters, including the learning rate, identical and relying solely on the scaling analysis of \cite{liu2025muonscalable} to port them. 
Unlike the experiments in previous sections, we also do \emph{not} apply the Polar Express re-parameterization~\cite{amsel2025polarexpress}. 
Despite this minimal adaptation, Muon converges faster and sustains a lower loss throughout pre-training.
These gains transfer to fine-tuning: \Cref{fig:finetune_ablation_combined} (left) reports terminal Acc@1 across all combinations of PT and FT optimizers, revealing a consistent benefit when Muon is used throughout the pipeline. 
The right panel of \Cref{fig:finetune_ablation_combined} makes this trend clearer by plotting Acc@1 over FT epochs relative to the Muon--Muon baseline; the Muon--Muon configuration maintains a persistent advantage over all other optimizer pairings across training.

This experiment should be viewed as a lower bound on the possible performance gains provided by Muon in MAE training and fine-tuning. Further performance gains, as seen in all other experiments in this paper, are likely possible by using polar express coefficients \cite{amsel2025polarexpress} and tuning learning rates and other Muon hyperparameters for the self-supervised pretraining. 

\section{Conclusions and limitations}
We demonstrate that Muon consistently outperforms AdamW in training ViTs across classification, segmentation, and masked autoencoders, achieving particularly large gains on long-tailed Pl@ntNet macro top-1. 
Interestingly, Muon's success in classification is recipe-dependent, benefiting disproportionately from a ``full'' vision recipe that includes heavy augmentation, mixup, cutmix, and label smoothing. 
This performance gap is reflected in the gradient spectra: within Muon, removing strong data augmentation induces late-training mode concentration (primarily in deep MLP blocks), while under a fixed full recipe, AdamW's QKV gradients remain dominated by far fewer singular modes than Muon's. 
These two effects are complementary in the sense that the dominant effects of switching between AdamW and Muon, and Muon with and without a full augmentation recipe, arise in complementary subsets of matrix-valued parameters. 
Ultimately, Muon's advantage in ViTs is best understood as an optimizer--recipe interaction occurring across complementary parameter subsets. 
{Muon is most effective when the training recipe maintains rich matrix-gradient support for its orthogonalized updates to exploit.}

While our results demonstrate a clear advantage for Muon, our study has limitations and natural directions for future work. 
First, our recipe-ablation studies and spectral analyses are limited to ImageNet-100 and Pl@ntNet-300K; validating the robustness of the late-training spectral collapse phenomenon at larger scale remains an important next step. In the self-supervised domain, our foundation model experiment confirms Muon's potential on reconstruction objectives via MAE, even without any tuning specific to Muon. Tuning Muon hyperparameters as well as evaluating Muon's performance and gradient geometry under other frameworks, such as DINO, would provide a more complete picture. Last, we have made moderate efforts to tune AdamW and been consistently unable to match Muon performance, but acknowledge it is still feasible other hyperparameter settings could improve AdamW performance.

\subsubsection*{Acknowledgments and Disclosure of Funding}
This paper describes objective technical results and analysis. Any subjective views or opinions that might be expressed in the paper do not necessarily represent the views of the U.S. Department of Energy or the United States Government.

This article has been authored by an employee of National Technology \& Engineering Solutions of Sandia, LLC under Contract No. DE-NA0003525 with the U.S. Department of Energy (DOE). The employee owns all right, title and interest in and to the article and is solely responsible for its contents. The United States Government retains and the publisher, by accepting the article for publication, acknowledges that the United States Government retains a non-exclusive, paid-up, irrevocable, world-wide license to publish or reproduce the published form of this article or allow others to do so, for United States Government purposes. The DOE will provide public access to these results of federally sponsored research in accordance with the DOE Public Access Plan \url{https://www.energy.gov/downloads/doe-public-access-plan}. The work performed at Sandia National Laboratories was supported by the U.S. Department of Energy, Office of Science, Office of Advanced Scientific Computing Research, DyGenAI project, and the SEA-CROGS project in the MMICCs program. Additional support was received from Interlab 
 Laboratory Directed Research and Development program at Sandia.

This work was funded in part by the National Nuclear Security Administration Interlab Laboratory
Directed Research and Development program under project number 20250861ER. This paper de-
scribes objective technical results and analysis. Any subjective views or opinions that might be
expressed in the paper do not necessarily represent the views of the U.S. Department of Energy or
the United States Government. Sandia National Laboratories is a multimission laboratory managed
and operated by National Technology and Engineering Solutions of Sandia, LLC., a wholly owned
subsidiary of Honeywell International, Inc., for the U.S. Department of Energy’s National Nuclear
Security Administration under contract DE-NA-0003525. SAND2026-21053O. The research was
performed under the auspices of the National Nuclear Security Administration of the U.S. Department
of Energy at Los Alamos National Laboratory, managed by Triad National Security, LLC under
contract 89233218CNA000001. Los Alamos National Laboratory Report LA-UR-25-30571.

\bibliographystyle{plain}
\bibliography{refs}

%%%%%%%%%%%%%%%%%%%%%%%%%%%%%%%%%%%%%%%%%%%%%%%%%%%%%%%%%%%%

\appendix
\clearpage

\section{Computational cost of Muon}\label{app:cost}

From a practical perspective, Muon is computationally more expensive than AdamW due to the approximate orthogonalization. Exactly how expensive relative to the full training pipeline depends on a variety of aspects like compute node, network architecture, and data set (e.g., \cite[Tabs. 3-5]{southworth2026beyond}). In our vision setting, we have found the overhead to be modest enough (and generally smaller than in language models) that the accuracy gains provided by Muon are significant when normalized in wall-clock terms. Table~\ref{tab:throughput} reports median training throughput on a single H100 for image classification using a ViT-B/16 architecture under the full recipe for AdamW, Muon, and MUD, a recent cheaper approximate orthogonalization method \cite{southworth2026beyond}. Muon is slower than AdamW on both ImageNet-100 and Pl@ntNet-300K, but the gap is smaller than is often assumed for matrix-aware optimizers. At the same time, cheaper alternatives such as MUD do not provide a sufficiently strong throughput advantage to offset their weaker optimizer performance in the ViT setting.

\begin{table}[!ht]
\centering
\small
\setlength{\tabcolsep}{5pt}
\begin{tabular}{lcccc}
\toprule
Optimizer & IN100 imgs/s & Rel. & PlantNet imgs/s & Rel. \\
\midrule
AdamW & \textbf{2557} & 1.00$\times$ & 2247 & 1.00$\times$ \\
MUD   & 2241 & 0.88$\times$ & \textbf{2251} & 1.00$\times$ \\
Muon  & 1980 & 0.77$\times$ & 2020 & 0.90$\times$ \\
\bottomrule\\
\end{tabular}
\caption{Mean training throughput (images/s) on a single H100 for ViT-B/16 under the full recipe. Muon is slower than AdamW, but the overhead is modest enough that its accuracy gains are practically relevant. MUD does not provide a compelling throughput advantage in the vision setting.}
\label{tab:throughput}
% \vspace{-3ex}
\end{table}

\section{Muon from a Whitening Perspective}
\label{app:whitening}

\subsection{Whitening, anisotropy suppression, and matrix updates}
\label{app:whitening_general}

Muon is typically presented as an approximate polar-factor update on matrix-valued gradients, but it is also useful to place Muon in a broader whitening perspective. Orthogonalized or whitened optimization updates attempt to remove anisotropy in a structured matrix geometry, replacing a raw momentum or gradient matrix with a transformed update that is less redundant and better conditioned. Muon is the canonical recent example of this idea \cite{jordan2024muon,liu2025muonscalable}, replacing a momentum matrix update with an approximation to its polar factor. This falls into a broader family of optimizers built on orthogonalization or whitening transformations. Here we discuss the broader family, potential algorithmic realizations, and discuss variations we have tried in the context of ViTs. Our eventual conclusion is that in the context of optimizers built on approximate whitening for ViTs, standard Muon and modifications thereof that approximate the polar factor appear to be the most effective in terms of accuracy per computational cost. 

If $M \in \mathbb{R}^{k \times d}$ is a matrix update and
$G = M M^\top,$
then any left transform $W$ satisfying
\[
W G W^\top = I
\]
produces a whitened representation $\widetilde M = W M$. In general there are infinitely many such transforms, differing by an arbitrary orthogonal rotation, but there are five canonical whitening transformations obtained by selecting natural optimality criteria \cite{kessy2018optimal}. In the matrix-valued setting, these five canonical choices can be written as follows:
\begin{enumerate}
    \item \textbf{ZCA whitening} -- The ZCA transform is
    $W_{\mathrm{ZCA}} = G^{-1/2} \implies \widetilde M_{\mathrm{ZCA}} = G^{-1/2} M.$ If $M = U \Sigma V^\top \implies G = U\Sigma^2 U^T$ is the thin singular value decomposition, then $\widetilde M_{\mathrm{ZCA}} = U V^\top,$ which is exactly the left polar factor of $M$. Thus, in our setting, ZCA whitening and polar orthogonalization coincide \cite{kessy2018optimal,lau2025polargrad}.

    \item \textbf{PCA whitening} -- PCA whitening applies the inverse square root in the eigenbasis of $G$:
    $\widetilde M_{\mathrm{PCA}} = \Sigma^{-1} U^\top M = V^\top$. This is expressed in principal coordinates rather than the original row basis \cite{kessy2018optimal}. Unlike ZCA, PCA whitening requires explicit access to the eigenbasis $U$ and therefore does not admit a simple Muon-style polynomial iteration in terms of $G$ alone.

    \item \textbf{Cholesky whitening} -- If $G = C C^\top$ is the Cholesky factorization, then $\widetilde M_{\mathrm{chol}} = C^{-1} M.$ This whitening transformation is asymmetric and ordering-dependent \cite{kessy2018optimal}, but the factorization is deterministic rather than iterative in nature (as singular value or eigendecompositions are).

    \item \textbf{ZCA-cor whitening} --
    Let $D = \mathrm{diag}(G)$, or equivalently $D_{ii} = \|M_{i:}\|^2$, and $P = D^{-1/2} G D^{-1/2}.$
    Then ZCA-cor whitening is $\widetilde M_{\mathrm{ZCAcor}} = P^{-1/2} D^{-1/2} M.$
    Equivalently, one first row-normalizes $M$ and then applies ZCA whitening. In our matrix-update language, this is a \emph{row-normalized polar factor}, and can be approximated in analogous ways as Muon.
    
    \item \textbf{PCA-cor whitening} --
    Similarly, let $\widetilde M_{\mathrm{PCAcor}} = \Theta^{-1/2} H^\top D^{-1/2} M,$
    where $P = H \Theta H^\top$ is the eigendecomposition of the correlation matrix. As with PCA whitening, this method requires an explicit eigenspace rotation, which cannot be approximated via cheap polynomial iterations.
\end{enumerate}

As stated above, PCA and PCA-cor require explicit eigenvector extraction and thus do not admit simple Muon-style polynomial approximations, making them impractical for transformer optimization.
Cholesky whitening is the natural triangular option and is closely connected to MUD \cite{southworth2026beyond}. However, our experience in LLMs and ViTs is that MUD and even exact Cholesky whitening lead to worse optimization performance than Muon-style polar approximations. In the context of LLMs, the reduced overhead of MUD can yield significant improvements in validation loss with respect to wallclock time. For ViTs, we have not found any settings in which MUD or Cholesky wins in terms of loss with respect to time. 

Once the target is identified as the polar factor, the next question is how to approximate it efficiently. The original Muon update uses a short Newton--Schulz polynomial iteration, which is attractive because it requires only matrix multiplications and is therefore highly GPU-friendly \cite{jordan2024muon,liu2025muonscalable}. It is natural to try higher-quality iterative schemes such as Halley-type methods or Zolotarev rational approximations. We have tried a number of variations in each of these classes of methods, using Cholesky decompositions to compute shifted inverses. However, computational cost aside, we have seen comparable or at best marginal improvements over Muon with polar express coefficients, but not enough to justify their higher computational cost and wallclock times. Finally, we also considered ZCA-Corr-inspired modifications that blend Muon-style normalization with MUD-like or correlation-normalized variants. Specifically, we incorporate a row normalization of $M$ before computing the Muon update. Across ImageNet-100 and Pl@ntNet-300K, this method tracked standard Muon very closely and did not produce a clear improvement, so we do not explore further in this work.

\subsection{A Linear Model Analysis: Muon, Augmentation, and Spectral Invariance}
\label{app:linear_model}
The whitening perspective makes precise an intuition about data augmentation that motivates using
Muon over gradient descent or Adam for vision transformers. We formalize this with a simple linear
model and derive an exact expression for the Muon update that makes the augmentation-invariance
explicit.

\paragraph{Setup.}
Consider a linear model $f(x) = Wx$ with $W \in \mathbb{R}^{m \times d}$, trained under MSE loss
on data $\{(x_i, y_i)\}_{i=1}^n$:
\[
    L = \frac{1}{2n} \sum_{i=1}^n \| Wx_i - y_i \|^2.
\]
Let $X = [x_1, \ldots, x_n] \in \mathbb{R}^{d \times n}$ and $\Sigma = \frac{1}{n} XX^\top \in
\mathbb{R}^{d \times d}$ denote the input covariance. Writing $E = W - W^*$ for the error relative
to the optimal weights, the gradient factors as
\begin{equation}
    G = \nabla_W L = \frac{1}{n}(WXX^\top - YX^\top) = E\Sigma.
    \label{eq:gradient_factored}
\end{equation}
For natural images, $\Sigma$ is approximately low-rank, reflecting the near-manifold structure of
the data. Gradient descent therefore exhibits spectral bias, preferentially reducing error along
the top eigendirections of $\Sigma$.

\paragraph{Effect of data augmentation.}
Let $\{A_k\}_{k=1}^K$ be a family of augmentation operators (random crops, color jitter, Gaussian
blur, etc.) acting linearly on the input. The augmented input covariance becomes
\begin{equation}
    \tilde{\Sigma} = \frac{1}{K} \sum_{k=1}^K A_k \Sigma A_k^\top.
    \label{eq:augmented_cov}
\end{equation}
Unlike orthogonal transformations, most augmentations are not
isometries. Their effect smear the spectrum of $\Sigma$ and so under augmentation,
the gradient becomes $G = E\tilde{\Sigma}$, and gradient descent convergence is governed by
$\tilde{\Sigma}$:
\[
    E_{t+1} = E_t(I - \eta \tilde{\Sigma}).
\]
Although augmentation enriches the training data, it simultaneously worsens the conditioning seen
by gradient-based optimizers, because $\tilde{\Sigma}$ has a less favorable spectrum than $\Sigma$.

\paragraph{Muon's spectral invariance.}
Recall that Muon's update is the polar factor of the gradient:
\[
    \mathrm{ONS}(G) = UV^\top, \quad G = U\Sigma_G V^\top,
\]
which can equivalently be written as $\mathrm{ONS}(G) = G(G^\top G)^{-1/2}$. Using the factored
gradient $G = E\tilde{\Sigma}$ and the symmetry of $\tilde{\Sigma}$, we have
\[
    G^\top G = \tilde{\Sigma} E^\top E \tilde{\Sigma}.
\]
At or near initialization, the error $E$ has no preferred direction, so it is natural to assume
$E^\top E \sim \alpha I$ for some scalar $\alpha > 0$. Under this isotropy assumption,
\[
    G^\top G \sim \alpha \tilde{\Sigma}^2,
\]
and therefore
\[
    \mathrm{ONS}(G) = E\tilde{\Sigma} \cdot (\alpha \tilde{\Sigma}^2)^{-1/2}
    = \frac{1}{\sqrt{\alpha}} E.
\]
The augmented covariance $\tilde{\Sigma}$ cancels exactly. The Muon update step thus becomes
\[
    E_{t+1} = E_t\!\left(1 - \frac{\eta}{\sqrt{\alpha}}\right),
\]
which decreases geometrically at a rate independent of $\tilde{\Sigma}$. In contrast, gradient
descent converges at a rate controlled by the smallest eigenvalue of $\tilde{\Sigma}$, which
degrades as augmentation spreads the spectrum.

This is a concrete instance of the spectral anisotropy suppression described in
\cref{app:whitening_general}. 
By collapsing the singular values of $G$, Muon implicitly
preconditions by $\tilde{\Sigma}^{-1}$, inverting the distortion introduced by augmentation and
rendering the optimizer invariant to it. The benefit of augmentation is thus obtained without the usual optimization penalty.

\paragraph{Remark.}
The isotropy assumption $E^\top E \sim \alpha I$ is reasonable near initialization but breaks
down later in training as the model fits certain directions faster than others. The analysis
therefore applies most cleanly to the early-training regime. It nonetheless provides a reasonable heuristic for why Muon's polar-factor update is particularly well matched to vision transformer
training under heavy augmentation, consistent with the empirical results in the main text.

\section{ImageNet-100 and Pl@ntNet-300K}
\label{app:plantnet}

\subsection{Optimizer and recipe hyperparameters}
\label{app:plantnet:params}

Unless otherwise noted, all main experiments in \Cref{sec:main_results}-\Cref{sec:muon_adamw} train ViT-B/16 with batch size 512. We use cosine learning-rate decay with a minimum learning-rate ratio of $0.05$, which we found to perform best among the schedules tested. For AdamW, we tested learning rates $3\times 10^{-4}$, $4.5\times 10^{-4}$, $6\times 10^{-4}$, and $10^{-3}$. Among these, $3\times 10^{-4}$ gave the best combination of stability and final validation accuracy for ViT training, and is therefore used in the main experiments. Interestingly, this preference for a smaller AdamW learning rate appeared specific to the ViT setting: in separate Xception-style convolutional experiments, AdamW performed well at $10^{-3}$, as did Muon. For Muon, we found $10^{-3}$ to work best overall. Smaller values were consistently slower to reduce loss, while larger values such as $2\times 10^{-3}$ showed mild instability or slower convergence.

Mixup/CutMix and stronger augmentation are either enabled at their default values or disabled entirely, depending on the recipe variant. The \emph{Full} recipe uses the default mixing and augmentation settings from our training code. The \emph{No Rand} variant disables RandAugment and random erasing while retaining Mixup/CutMix. The \emph{No Mix} variant disables Mixup/CutMix while retaining stronger image augmentation. The \emph{No Mix/No Rand} variant disables both mixing-side smoothing and stronger augmentation. 

All training runs use baseline spatial preprocessing consisting of random resized cropping with minimum scale $0.08$ and random horizontal flipping with probability $0.5$, while evaluation uses crop ratio $0.875$. On top of this baseline, the \emph{Full} recipe adds two components. First, \emph{strong augmentation} consists of RandAugment with 2 sampled operations at magnitude 9 together with Random Erasing at probability $0.25$. Second, \emph{mixing-side smoothing} consists of Mixup with $\alpha=0.8$, CutMix with $\alpha=1.0$, mixing probability $1.0$, Mixup/CutMix switch probability $0.5$, and label smoothing $0.1$. In our experiments, Mixup, CutMix, and label smoothing are enabled or disabled together, so we treat them as a single component in the main text. These components can be implemented directly with standard Torchvision transforms for RandAugment, Random Erasing, MixUp, and CutMix, while label smoothing is applied in the loss rather than as an image transform.

All other optimizer and model settings are kept fixed across the main AdamW--Muon comparisons unless otherwise stated.

\begin{table}[t]
\centering
\small
\setlength{\tabcolsep}{6pt}
\begin{tabular}{ll}
\toprule
Setting & Value \\
\midrule
Model & ViT-B/16 \\
Main batch size & 512 \\
Appendix batch size & 128 \\
Schedule & Cosine decay \\
Minimum LR ratio & 0.05 \\
AdamW LR sweep & $3\mathrm{e}{-4},\,4.5\mathrm{e}{-4},\,6\mathrm{e}{-4},\,1\mathrm{e}{-3}$ \\
AdamW LR used & $3\mathrm{e}{-4}$ \\
Muon LR sweep & $6\mathrm{e}{-4},\,1\mathrm{e}{-3},\,2\mathrm{e}{-3}$ \\
Muon LR used & $1\mathrm{e}{-3}$ \\
Recipes & Full / No Rand / No Mix / No Mix/No Rand \\
\bottomrule
\end{tabular}
\caption{Main optimizer and recipe hyperparameters used in the paper.}
\label{tab:hyperparams}
\end{table}

\subsection{Batch size 128}
\label{app:bsize128}

\begin{figure}[t]
    \centering
    \includegraphics[width=0.9\columnwidth]{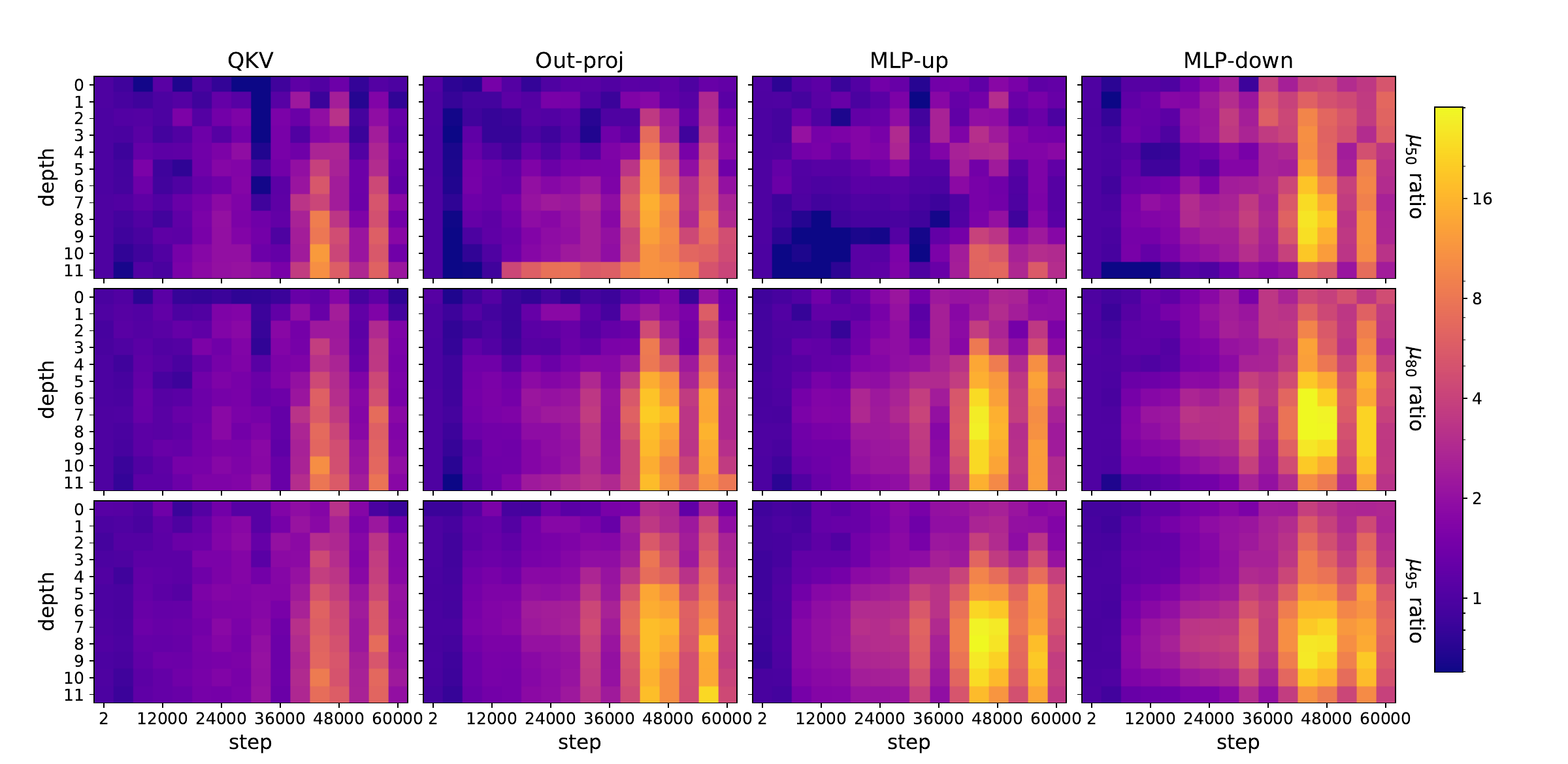}
    \caption{Batch size 128 energy-quantile rank ratio for Muon trained with the \emph{Full} recipe over Muon trained with \emph{No Mix/No Rand} for gradient matrices $G$. Values greater than one indicate broader spectra under the full recipe. Results are largely analogous to batch size 512 in \Cref{fig:muon_aug_atlas}.}
    \label{fig:muon_aug_atlas128}
    % \vspace{-4ex} % looks weird 
\end{figure}

We tested Muon with the \emph{Full} recipe and \emph{No Mix/No Rand} on batch size 128 as well, to see if reduced batch size affects performance and/or gradient spectra behavior, in particular due to the direct relation between matrix-parameter achievable rank and batch size. After processing the same number of images (60,000 steps here), final validation results for Muon are shown in \Cref{tab:muon-val-plantnet}. Results for batch size 128 are largely analogous, with slightly improved \emph{Full} recipe final performance and larger performance gap to \emph{No Mix/No Rand} recipe on the Macro Top-1. Cumulative energy-quantile rank ratio for Muon trained with the \emph{Full} recipe over Muon trained with \emph{No Mix/No Rand} are shown in \Cref{fig:muon_aug_atlas128}; results largely match the batch size 512 case in \Cref{fig:muon_aug_atlas}, with slightly larger ratios seen here for batch size 128.

\begin{table}[!ht]
    \centering
    \caption{Final validation metrics on Pl@ntNet-300K.}
    \label{tab:muon-val-plantnet}
    \begin{tabular}{lc | ccc | ccc}
        \toprule
        & & \multicolumn{3}{c|}{Batch size 128}
        & \multicolumn{3}{c}{Batch size 512} \\
    Optimizer & Recipe & Loss & Top-1 & Macro & Loss & Top-1 & Macro \\
    \midrule
    Muon & Full           & 0.88 & 80.95 & 37.34 & 0.91 & 80.26 & 37.08 \\
    Muon & No Mix/No Rand & 1.40 & 73.89 & 30.33 & 1.42 & 73.89 & 31.27 \\
        \bottomrule
    \end{tabular}
\end{table}

\subsection{Spectral analysis}
\label{app:spectral_details}

\begin{figure}[!ht]
    \centering
    \includegraphics[width=0.9\columnwidth]{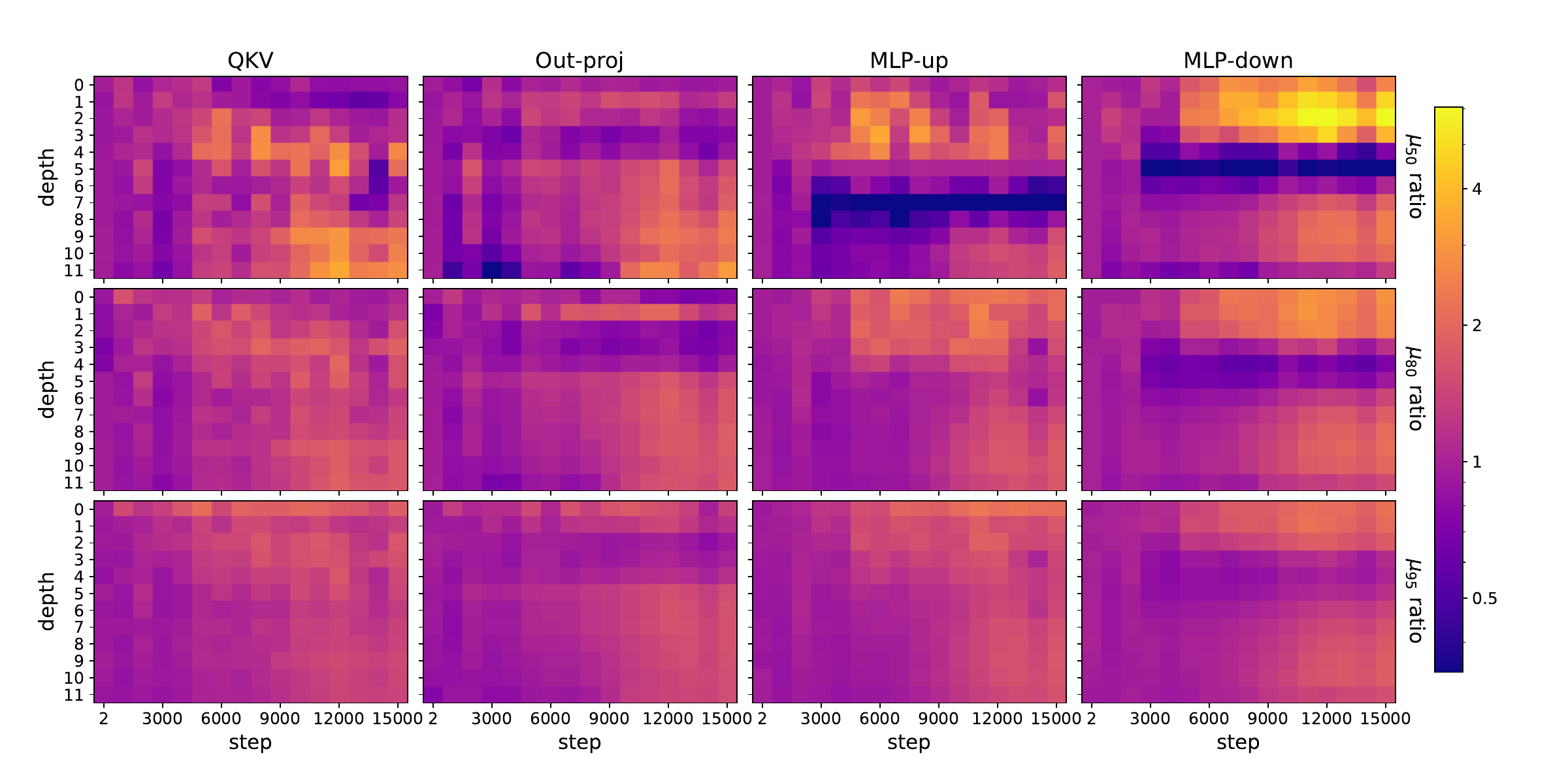}
    \caption{Energy-quantile rank ratio for Muon trained with the full recipe over Muon trained with \emph{No Mix/No Rand} for gradient matrices $G$. Values greater than one indicate broader spectra under the full recipe. The strongest late-training effect is concentrated in deep MLP-down blocks.}
    \label{fig:muon_aug_atlasM}
    % \vspace{-4ex} % looks weird 
\end{figure}

Analogous cumulative quartile energy ratios as in \Cref{fig:muon_aug_atlas} are shown in \Cref{fig:muon_aug_atlasM} for momentum matrices. Here, we see there is much less structure and differences between a \emph{Full} recipe and \emph{No Mix/No Rand} recipe. 

\begin{figure}[!ht]
    \centering
    \includegraphics[width=\columnwidth]{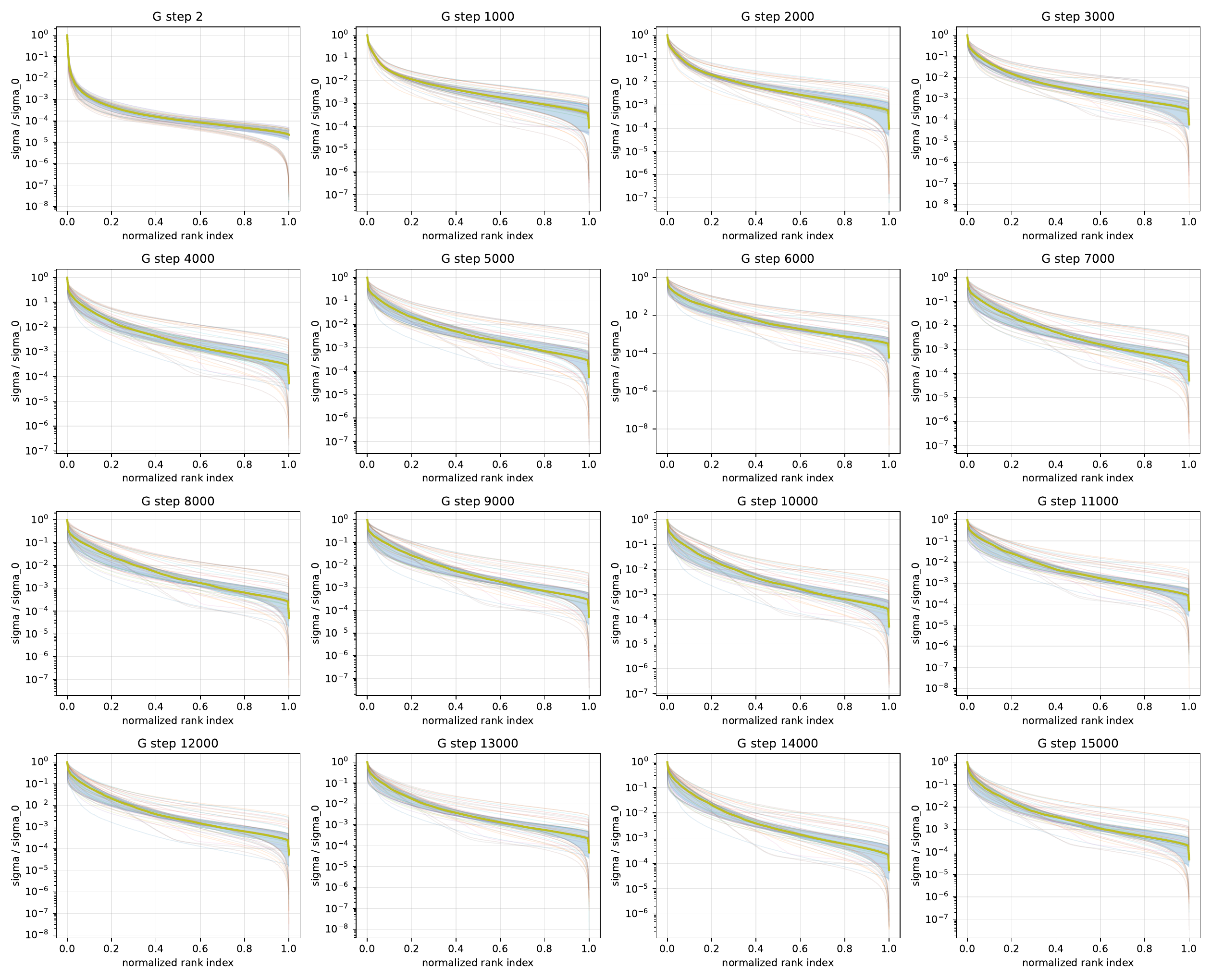}
    \caption{Normalized singular values across all gradient weight matrices in architecture for AdamW with \emph{No Rand/No Mix} training recipe.}
    \label{fig:spectrum_adamw}
    % \vspace{-4ex} % looks weird 
\end{figure}

\begin{figure}[!ht]
    \centering
    \includegraphics[width=\columnwidth]{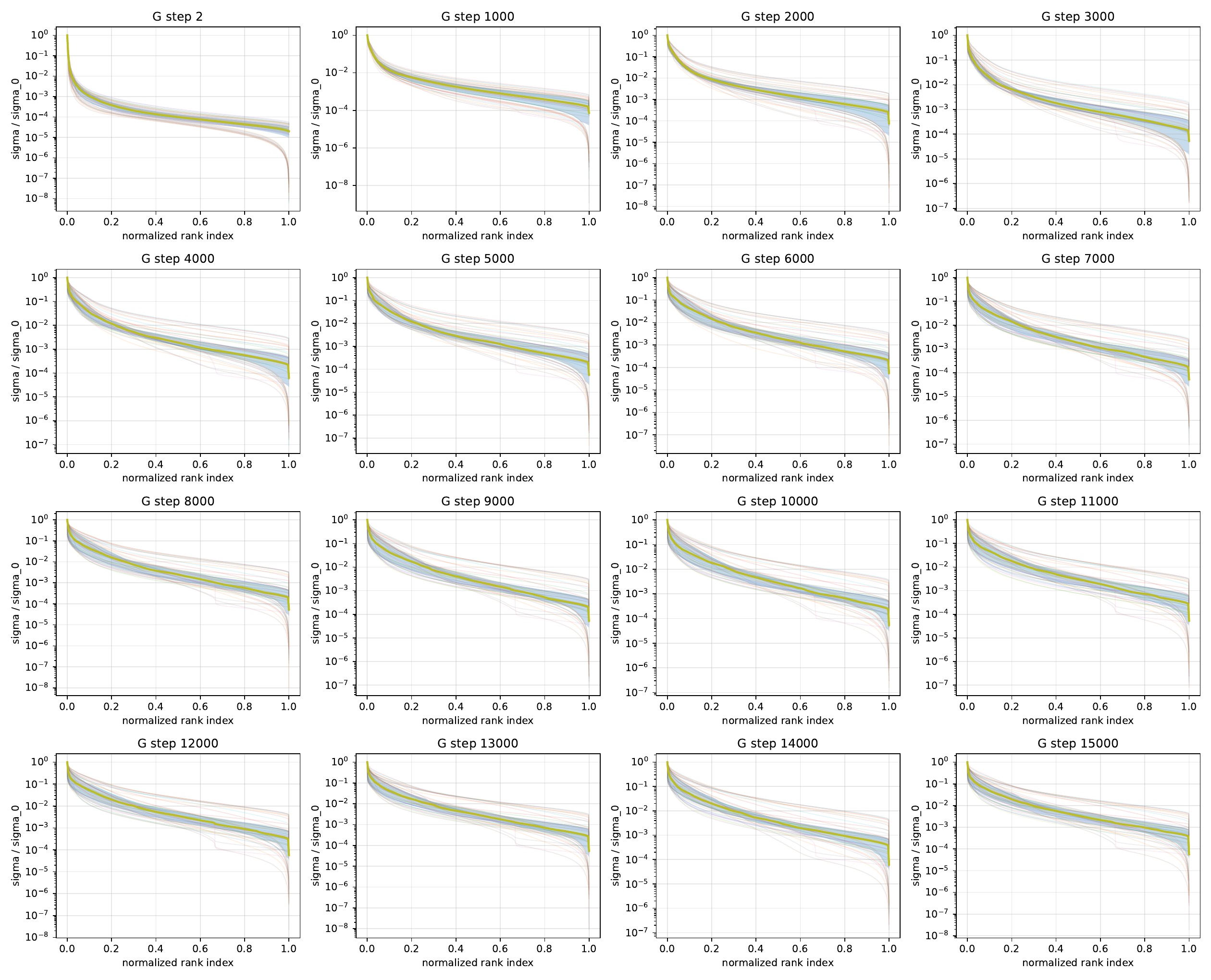}
    \caption{Normalized singular values across all gradient weight matrices in architecture for AdamW with \emph{Full} training recipe.}
    \label{fig:spectrum_adamw_full}
    % \vspace{-4ex} % looks weird 
\end{figure}

\begin{figure}[!ht]
    \centering
    \includegraphics[width=\columnwidth]{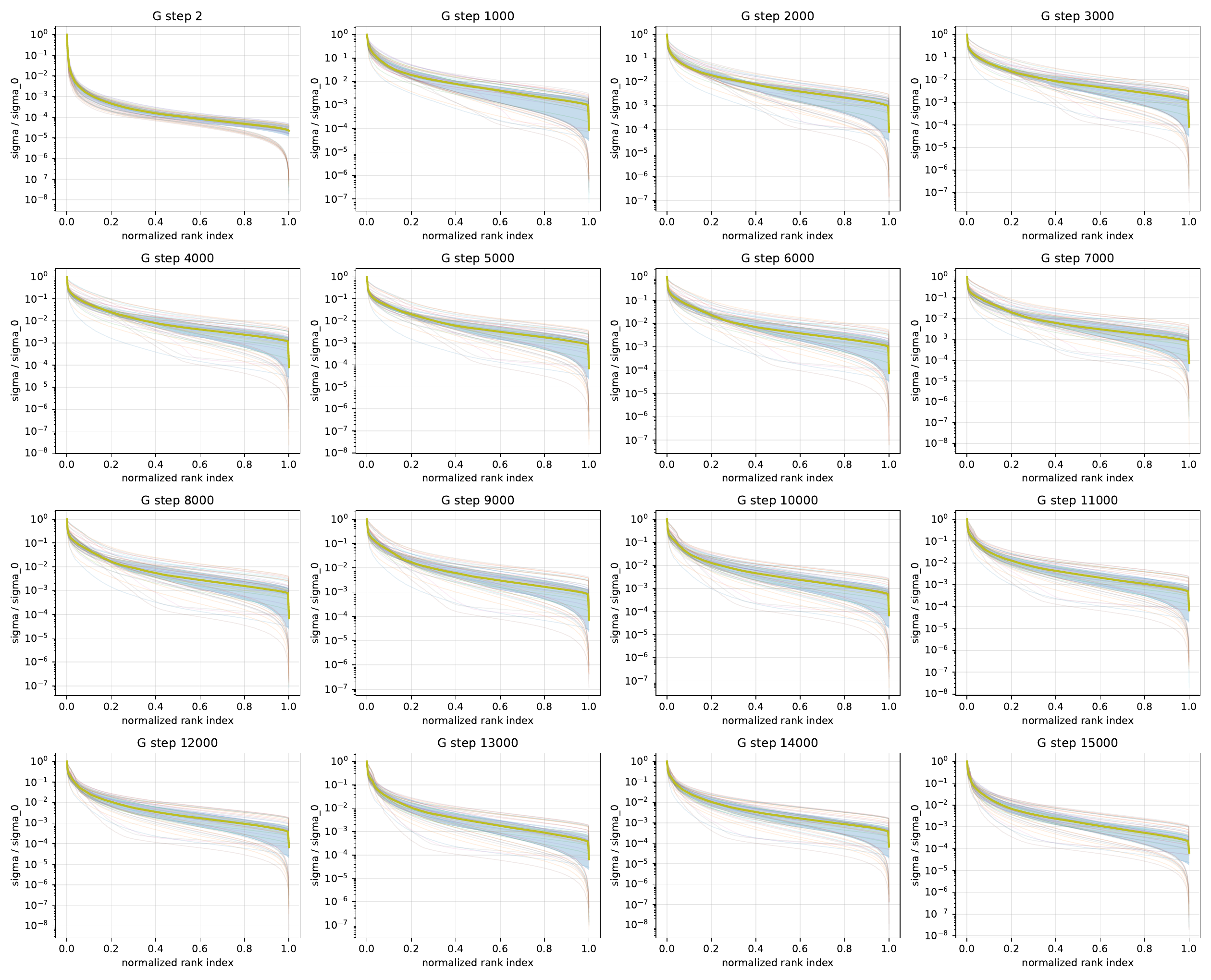}
    \caption{Normalized singular values across all gradient weight matrices in architecture for Muon with \emph{No Rand/No Mix} training recipe.}
    \label{fig:spectrum_muon}
    % \vspace{-4ex} % looks weird 
\end{figure}

\begin{figure}[!ht]
    \centering
    \includegraphics[width=\columnwidth]{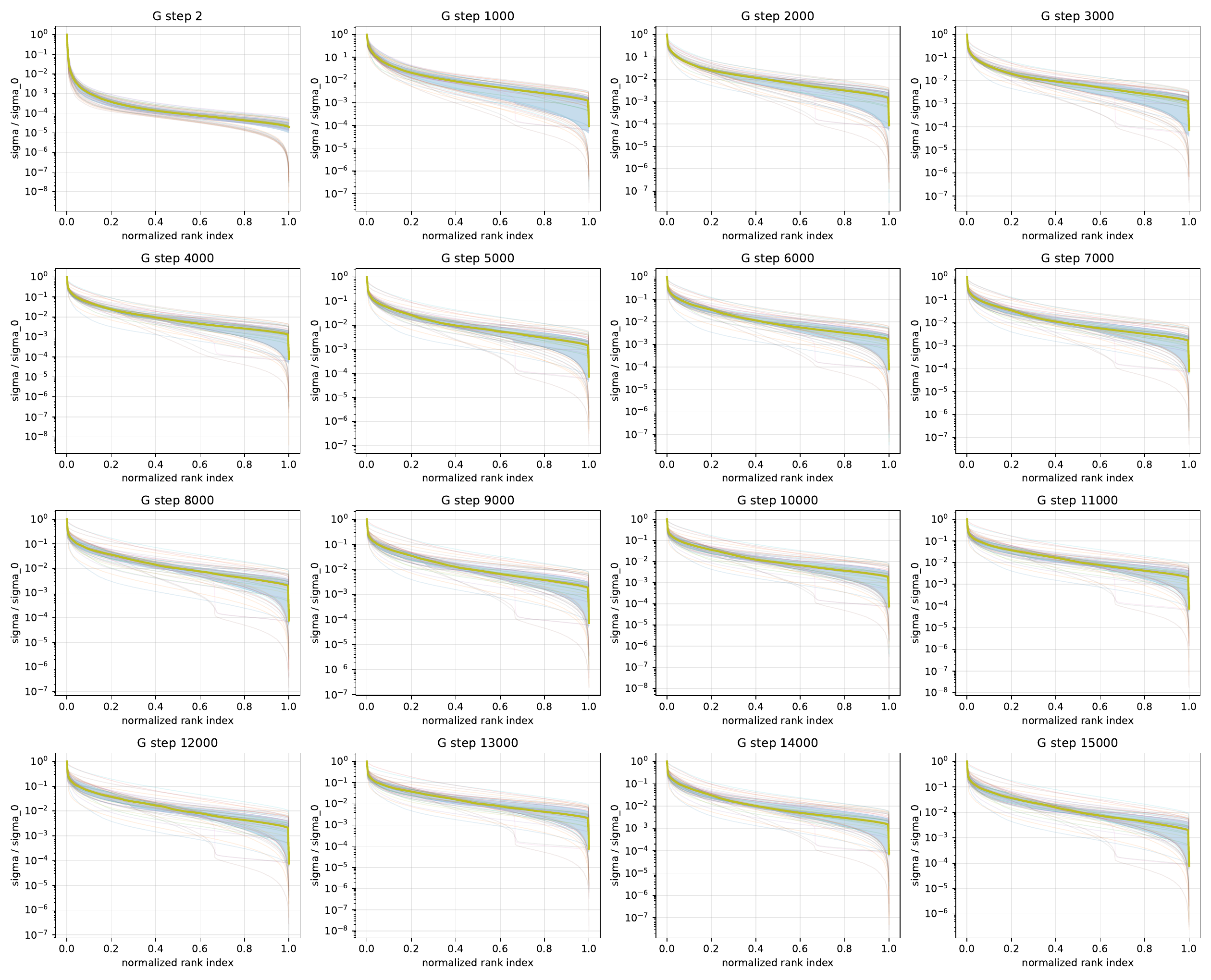}
    \caption{Normalized singular values across all gradient weight matrices in architecture for Muon with \emph{Full} training recipe.}
    \label{fig:spectrum_muon_full}
    % \vspace{-4ex} % looks weird 
\end{figure}

\begin{figure}[!ht]
    \centering
    \includegraphics[width=\columnwidth]{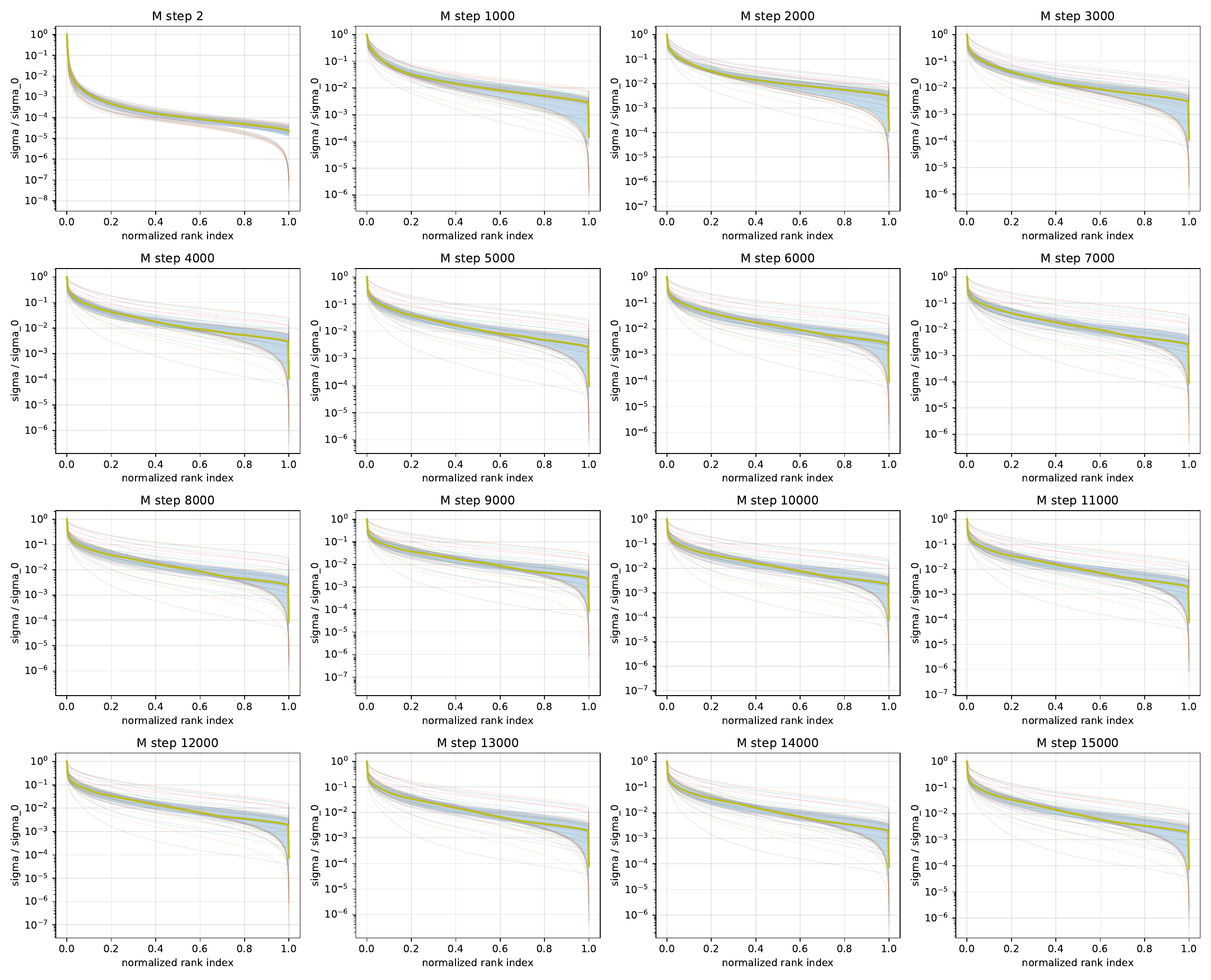}
    \caption{Normalized singular values across all momentum weight matrices in architecture for Muon with \emph{No Rand/No Mix} training recipe.}
    \label{fig:spectrum_muonM}
    % \vspace{-4ex} % looks weird 
\end{figure}

\begin{figure}[!ht]
    \centering
    \includegraphics[width=\columnwidth]{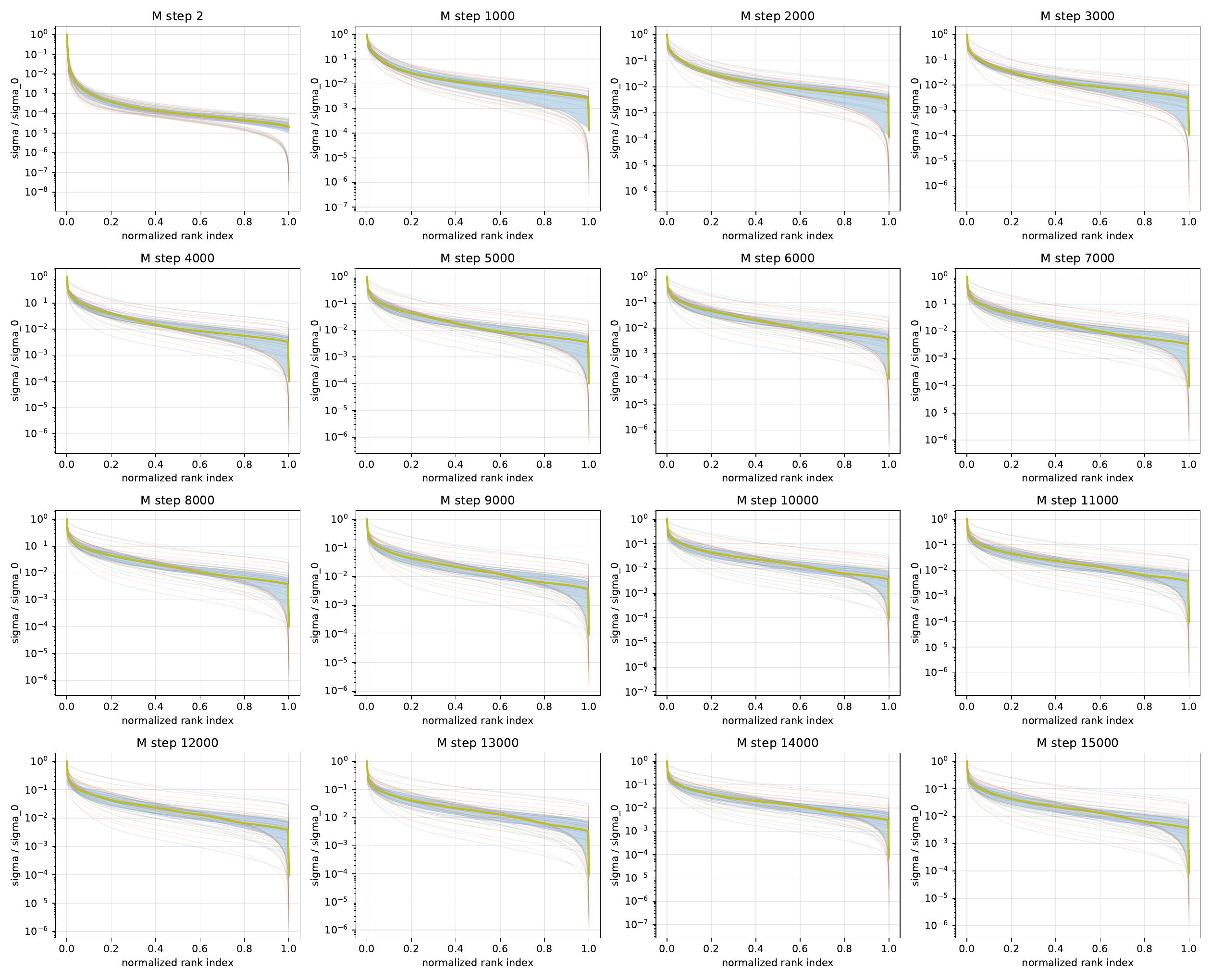}
    \caption{Normalized singular values across all momentum weight matrices in architecture for Muon with \emph{Full} training recipe.}
    \label{fig:spectrum_muonM_full}
    % \vspace{-4ex} % looks weird 
\end{figure}

Identifying the cumulative energy as a clear metric distinguishing optimization trajectories and gradient behavior between different optimizers and training recipes originated by considering the evolving singular value distribution of matrix-valued parameters. \Cref{fig:spectrum_adamw}-\Cref{fig:spectrum_muonM_full} provide plots of the spectrum across all linear operators, from initial to final training step. Slower spectral decay for Muon than AdamW, and for Muon with a \emph{Full} recipe compared with \emph{No Mix/No Rand} can be observed in these plots as well.


\subsection{Xception on ImageNet-100}
\label{app:plantnet:xception}

To test whether the main empirical pattern extends beyond transformers, we also trained an Xception \cite{chollet2017xception} convolutional neural network on ImageNet-100 under the \emph{No Mix/No Rand} and \emph{Full} recipes. For the Xception experiments, we use the same hybrid optimizer framework as usual, but Muon is applied only to convolution weight tensors with kernel size $1\times 1$ and more than one output channel, including the pointwise projections inside each depthwise-separable convolution block and the $1\times 1$ skip-path projection convolutions used when the channel dimension or stride changes. 
A $1\times 1$ convolution is a shared channel-mixing linear map: at each spatial location it applies the same matrix $W \in \mathbb{R}^{C_{\mathrm{out}}\times C_{\mathrm{in}}}$ to the local channel vector. For this reason, we treat these pointwise convolutions as the convolutional layers most directly analogous to the dense learned linear maps in a ViT. We do not include the spatial $3\times 3$ stem convolutions, the depthwise convolution kernels, normalization parameters, biases, or the final fully connected classifier, which are all optimized by the AdamW branch of the hybrid optimizer. Thus, in Xception, Muon acts only on the subset of layers that are most naturally interpreted as matrix-valued channel-mixing operators. This differs from the ViT setting, where a much larger fraction of the architecture is composed of dense learned linear maps.

The results are summarized in Table~\ref{tab:xception_appendix}. The same qualitative recipe effect is present: both AdamW and Muon improve under the full recipe, and the gain is slightly larger for Muon. However, the overall Muon--AdamW gap is much smaller than for ViT-B/16. The weak-recipe setting is also informative. Under \emph{No Mix/No Rand}, Muon achieves substantially higher training accuracy than AdamW but slightly worse validation accuracy, indicating more severe overfitting. Under the \emph{Full} recipe, this gap in training behavior largely disappears and Muon slightly outperforms AdamW on validation top-1. This suggests that outside the transformer setting, Muon may overfit more aggressively when the recipe is weak, and that the full vision recipe is important for controlling this behavior. It is possible imposing additional matrix structure in Xception and CNN architectures beyond just the $1\times 1$ convolutions could help Muon perform better, but this is outside the scope of this study.  

\begin{table}[t]
\centering
\small
\setlength{\tabcolsep}{5pt}
\begin{tabular}{llcccc}
\toprule
Model & Optimizer & Recipe & Train Top-1 & Val Top-1 & Val Top-5 \\
\midrule
Xception & AdamW & No Mix/No Rand & 93.95 & 83.00 & 95.44 \\
Xception & Muon  & No Mix/No Rand & 97.27 & 82.40 & 94.82 \\
Xception & AdamW & Full           & --- & 85.80 & 97.02 \\
Xception & Muon  & Full           &  --- & 86.36 & 96.74 \\
\bottomrule
\end{tabular}
\caption{ImageNet-100 results for Xception under \emph{No Mix/No Rand} and the \emph{Full} recipe. Under the weak recipe, Muon attains higher training accuracy but slightly worse validation accuracy than AdamW, indicating stronger overfitting. Under the full recipe, this overfitting is mitigated and Muon slightly outperforms AdamW on validation top-1. Training top-1 is not included for the Full recipe, as it is strongly affected by the mixing-side smoothing and data augmentation, and not representative of model accuracy.}
\label{tab:xception_appendix}
\end{table}

\section{Image segmentation task specification and supplemental analysis}
\label{app:image_segmentation}

In this section we specify experimental settings and present supplemental analysis for the image segmentation experiments.

\subsection{Experimental specification}

The image segmentation study uses LoveDA semantic segmentation with a compact SegFormer model trained from scratch. 
See \Cref{tab:seg-appendix-spec} for a specific details. 
We compare AdamW and Muon under four augmentation regimes. The augmentation regimes are ordered by increasing augmentation strength as \texttt{noaug}, \texttt{base}, \texttt{geo}, and \texttt{geophoto}. Across all runs, the architecture, training horizon, crop size, and general optimization setup are held fixed so that the main intended differences are optimizer, augmentation regime, and seed.

\begin{table}[h]
    \centering
    \caption{Compact specification of the LoveDA image segmentation experiment.}
    \label{tab:seg-appendix-spec}
    \begin{tabular}{ll}
        \toprule
        Item & Specification \\
        \midrule
        Task & Semantic segmentation \\
        Dataset & LoveDA \\
        Semantic classes & 7 \\
        Training split & Urban: 1156 images; Rural: 1366 images \\
        Validation split & Urban: 677 images; Rural: 992 images \\
        Loss & Cross Entropy \\
        Model & Compact SegFormer \\
        Encoder stages & 4 \\
        Embedding dimensions & \([32, 64, 160, 256]\) \\
        Depths & \([2, 2, 2, 2]\) \\
        Attention heads & \([1, 2, 5, 8]\) \\
        MLP ratios & \([4, 4, 4, 4]\) \\
        Spatial-reduction ratios & \([8, 4, 2, 1]\) \\
        Decoder dimension & \(128\) \\
        Approx.\ parameter count & \(\sim 3.45\)M \\
        Optimizers & AdamW, Muon \\
        Augmentation regimes & \texttt{noaug}, \texttt{base}, \texttt{geo}, \texttt{geophoto} \\
        Training crop size & \(512 \times 512\) \\
        Training horizon & 3000 optimization steps \\
    \end{tabular}
\end{table}

The primary validation metric is mean intersection-over-union (mIoU), with overall accuracy (OA) and Cohen's kappa reported as supporting metrics.

\subsection{Augmentation regimes}

The four augmentation regimes are intended to form a controlled ladder from minimal task-safe preprocessing to heavier geometric and photometric perturbation. In all cases, the training pipeline includes a \(512 \times 512\) random crop followed by normalization and tensor conversion. The regimes differ only in the additional augmentation operators applied before normalization.

\begin{table}[h]
    \centering
    \caption{Augmentation regimes used in the LoveDA image segmentation study.}
    \label{tab:seg-appendix-augs}
    \begin{tabular}{p{0.16\textwidth}p{0.78\textwidth}}
        \toprule
        Regime & Specification \\
        \midrule
        \texttt{noaug} &
        RandomCrop\((512,512)\), Normalize, ToTensor. \\[2pt]

        \texttt{base} &
        RandomCrop\((512,512)\), OneOf\{HorizontalFlip, VerticalFlip, RandomRotate90\}, Normalize, ToTensor. \\[2pt]

        \texttt{geo} &
        \texttt{base} augmentation plus additional geometric perturbation via affine-style transforms. \\[2pt]

        \texttt{geophoto} &
        \texttt{geo} augmentation plus additional photometric perturbation from a color-transform pool (e.g., RGB shift / HSV / color jitter / gamma-style adjustments). \\
        \bottomrule
    \end{tabular}
\end{table}

\subsection{Supplemental analysis}

\begin{figure}[h]
    \centering
    \begin{subfigure}[t]{0.48\textwidth}
        \centering
        \includegraphics[width=\textwidth]{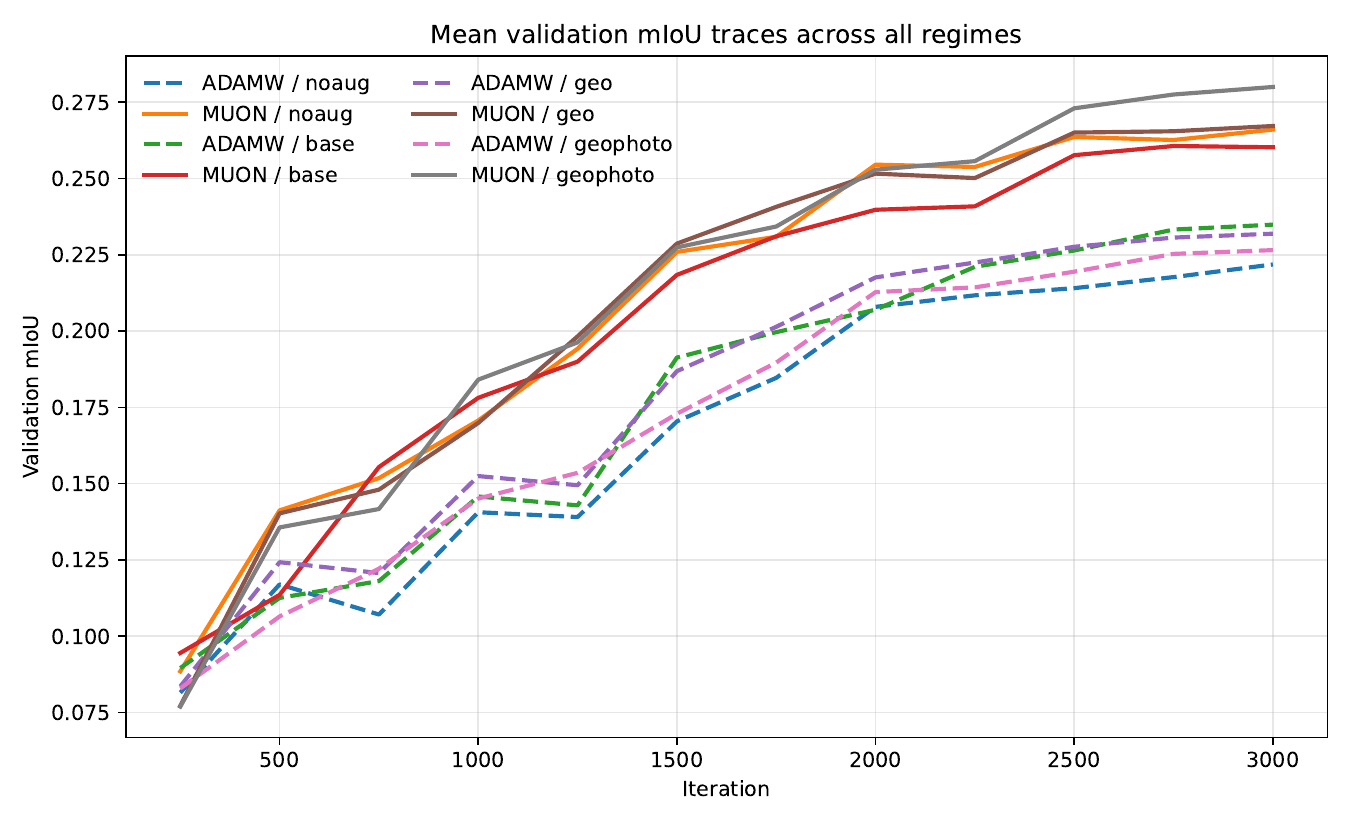}
        \caption{Mean validation mIoU traces across training for all optimizer--regime pairs. Each curve shows the mean over three seeds.}
        \label{fig:seg-appendix-miou-traces}
    \end{subfigure}
    \hfill
    \begin{subfigure}[t]{0.48\textwidth}
        \centering
        \includegraphics[width=\textwidth]{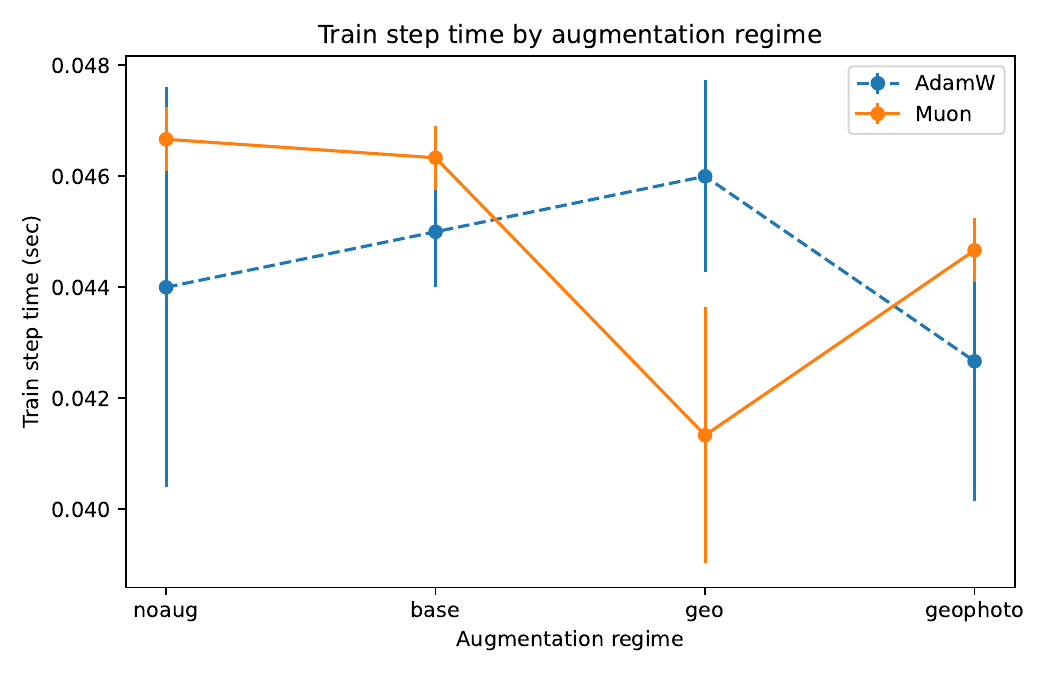}
        \caption{Mean training step time across augmentation regimes. Error bars denote standard deviation over three seeds.}
        \label{fig:seg-appendix-time}
    \end{subfigure}
    \caption{Supplemental segmentation results on LoveDA. Left: validation mIoU across training for all optimizer--regime pairs. Right: mean training step time by augmentation regime.}
    \label{fig:seg-appendix-miou-time}
\end{figure}

Here we collect additional analysis for the image segmentation experiments. Table~\ref{tab:seg-appendix-terminal-full} reports the same terminal validation results as in the main text, now augmented with uncertainty estimates (mean $\pm$ standard deviation over three seeds) for all three metrics. Figure~\ref{fig:seg-appendix-miou-traces} shows the validation mIoU trajectories across training for all optimizer--regime pairs. We also summarize training-time cost in Figure~\ref{fig:seg-appendix-time}.

\begin{table}[h]
    \centering
    \caption{Terminal validation results on LoveDA semantic segmentation for AdamW and Muon across augmentation regimes, reported as mean $\pm$ standard deviation over three seeds. Within each metric pair, the better mean value is shown in bold.}
    \label{tab:seg-appendix-terminal-full}
    \setlength{\tabcolsep}{4pt}
    \begin{tabular}{llcc}
        \toprule
        Regime & Metric & AdamW & Muon \\
        \midrule
        noaug     & mIoU  & 0.222 $\pm$ 0.011 & \textbf{0.266 $\pm$ 0.002} \\
        noaug     & OA    & 0.494 $\pm$ 0.010 & \textbf{0.522 $\pm$ 0.007} \\
        noaug     & Kappa & 0.303 $\pm$ 0.017 & \textbf{0.353 $\pm$ 0.006} \\
        \midrule
        base      & mIoU  & 0.235 $\pm$ 0.011 & \textbf{0.260 $\pm$ 0.010} \\
        base      & OA    & 0.498 $\pm$ 0.012 & \textbf{0.516 $\pm$ 0.008} \\
        base      & Kappa & 0.319 $\pm$ 0.014 & \textbf{0.348 $\pm$ 0.012} \\
        \midrule
        geo       & mIoU  & 0.232 $\pm$ 0.007 & \textbf{0.267 $\pm$ 0.003} \\
        geo       & OA    & 0.499 $\pm$ 0.006 & \textbf{0.526 $\pm$ 0.007} \\
        geo       & Kappa & 0.317 $\pm$ 0.012 & \textbf{0.360 $\pm$ 0.008} \\
        \midrule
        geophoto  & mIoU  & 0.227 $\pm$ 0.005 & \textbf{0.280 $\pm$ 0.004} \\
        geophoto  & OA    & 0.513 $\pm$ 0.008 & \textbf{0.556 $\pm$ 0.006} \\
        geophoto  & Kappa & 0.322 $\pm$ 0.008 & \textbf{0.385 $\pm$ 0.005} \\
        \bottomrule
    \end{tabular}
\end{table}

\section{Foundation Model Details}
\subsection{Masked Autoencoders}
Self-supervised pretraining methods learn visual representations without labeled data by defining auxiliary objectives that exploit the structure of images themselves, allowing for massive data consumption like classical text foundation models. 
MAE \cite{he2022masked} represent a paradigm within this space, and remain widely used reference point despite the subsequent development of more powerful methods.
The self-supervised pretraining landscape has since advanced considerably with DINO \cite{oquab2023dinov2,simeoni2025dinov3}, JEPA-based method \cite{assran2023self}, and others like data2vec \cite{baevski2022data2vec}. 
More recent foundation model pipelines increasingly combine self-supervised pretraining with large-scale supervised or vision--language objectives \cite{radford2021learning,jia2021scaling}, making clean optimizer comparisons substantially harder to isolate. 
For the purposes of studying the optimizer--recipe interaction introduced in the main paper, we therefore focus on the original MAE framework, which provide clean, well-understood pretraining objectives with minimal confounders.

MAE is a reconstruction-based method. 
During pretraining, a large fraction of image patches (75\% in our case) are randomly masked, and the model is trained to reconstruct the missing pixel content from the visible patches alone. 
The encoder processes only the visible tokens, while a lightweight decoder reconstructs the full image. 
The learned encoder representations are subsequently evaluated by fine-tuning on a downstream labeled dataset in the usual fashion for classification. 
Because the pretraining signal is entirely determined by which patches are masked and what must be reconstructed, augmentation in MAE operates at the level of the masking strategy itself.

% DINO is a self-distillation method based on a teacher--student framework. 
% A student network is trained to match the output distribution of a momentum-averaged teacher network across multiple augmented views of the same image. 
% The multi-crop strategy is where the student processes 
% several ``local'' crop views while the teacher processes ``global'' crops, encouraging the network to learn representations that are consistent across scale and viewpoint. 
% Other augmentations are also used such as strong color jitter, Gaussian blur, and 
% solarization are applied during view generation, making augmentation load-bearing at the level of the objective. 
% DINO representations are evaluated via $k$-NN classification and linear probing without any labeled fine-tuning of the backbone.

\subsection{Implementation Details}
For a fair comparison, we utilize the original GitHub repositories for MAE corresponding to \cite{he2022masked} located at \url{https://github.com/facebookresearch/mae}, and \emph{only} alter the optimizer, substituting AdamW with Muon for the backbone parameters while retaining AdamW for any remaining scalar or one-dimensional parameters (e.g., bias terms, LayerNorm scales) following standard Muon practice. 
All other hyperparameters during pretraining, evaluation or finetuning, including learning rate schedule, weight decay, batch size, number of epochs, and augmentation pipeline, are held fixed at the values specified in the respective original repositories. 
This ensures that any observed differences in downstream performance are attributable to the optimizer rather than to recipe or implementation differences.

We note that the learning rates in both repositories were originally tuned for AdamW. We make no attempt to retune them for Muon, and use the same schedules without modification. 
As stated in the main text, this is a conservative choice following \cite{liu2025muonscalable}, which is generally more optimized for LLMs. 
Choices in the Muon optimizer include values for the RMS update of 0.2 and Newton-Schulz iterations of 5, which can impact the convergence \cite{kim2026convergence}, which we never tuned but rather borrowed from the LLM environments. 
Dedicated learning rate and hyper-parameter tuning for Muon would likely yield further gains. 
The results reported here should therefore be understood as a {lower bound} on Muon's potential in these pretraining settings.

All MAE experiments were run using single NVIDIA A100~40\,GB GPUs for simplicity. 
Pre-training ablations consist of ViT-B pretrained for 500 epochs
and ViT-L pretrained for 400 epochs lasting approximately 250 GPU hours.
ImageNet fine-tuning ablations consisted of training for
50~epochs with an effective batch size of 1024 via gradient accumulation.
Linear probing ablations (see below) using 90 epochs for ViT-B,
50 for ViT-L at an effective batch size~16384 using at approximately 16~GPU-hours for the ViT-L.
Note again, that these are the default from the MAE repository.

\subsection{Linear Probe Result}
We also perform the basic linear probe of freezing the backbone and only training a linear classifier on Imagenet-1k; following the MAE repo, the linear classifier training is done using LARS.
The results are shown in \Cref{tab:mae-linprobe-terminal}, where it's clear that Muon maintains a significant advantage at roughly 2\% improvement over the AdamW pretrained model. 

  \begin{table}[t]                   
      \centering                                 
      \caption{Terminal validation results on ImageNet-1k MAE pre-training for AdamW and Muon across model sizes. Linear probing is run for 90 epochs (ViT-B) and 50 epochs (ViT-L) with lr\,=\,2.4e-3 using LARS, the default optimizer from the MAE repository. Within each metric pair, the better value is shown in bold.}  
      \label{tab:mae-linprobe-terminal}          
      \begin{tabular}{lcccc}                      
          \toprule                                            
          Model                                             
          & \multicolumn{2}{c}{Top-1 Acc.}                   
          & \multicolumn{2}{c}{Top-5 Acc.} \\                
          \cmidrule(lr){2-3}                                
          \cmidrule(lr){4-5}                                 
          & AdamW & Muon & AdamW & Muon \\                   
          \midrule                                            
          ViT-B & 60.71 & \textbf{62.06} & 82.67 & \textbf{83.69} \\  
          ViT-L & 67.29 & \textbf{69.65} & 87.05 & \textbf{88.67} \\  
          \bottomrule                                     
      \end{tabular}                            
  \end{table}

\end{document}